**Vision–Language Models for Occupational Physical Exposure Assessment: Estimating External Hand Forces in Manual Material Handling Tasks from RGB Video**

**Author names**:

[a] Mohammad Sadra Rajabi, https://orcid.org/0000-0002-9100-3973

[b] Aanuoluwapo Ojelade, https://orcid.org/0000-0001-9715-3254

[a] Sunwook Kim, https://orcid.org/0000-0003-3624-1781

[a] Maury A. Nussbaum, https://orcid.org/0000-0002-1887-8431

**Affiliations:**

[a] Department of Industrial and Systems Engineering, Virginia Tech, Blacksburg VA 24061, USA

[b] St. Jude Children's Research Hospital, Memphis, TN 38105, USA

**Corresponding author:**

Corresponding address: Maury A. Nussbaum

Department of Industrial and Systems Engineering,

Virginia Tech, Blacksburg VA 24061, USA.

Phone: 540-231-6053. Email: nussbaum@vt.edu

**Abstract**

External hand forces are important inputs to biomechanical analyses of occupational physical exposure and injury risk, yet continuous force measurements during manual material handling (MMH) typically requires instrumented objects or specialized sensing. We evaluated a vision–language model (VLM)-based pipeline that combines task-specific textual cues, visual representations, and known box mass to estimate dynamic, triaxial, bilateral external hand forces from RGB video. Thirty-five healthy young adults performed five MMH tasks involving lifting, carrying, pushing, and pulling with box masses of 6, 9, and 12 kg. The pipeline used text-guided localization of participant and handled-object regions of interest (ROIs), pretrained vision-transformer feature extraction, and transformer-based temporal regression. Performance was evaluated using leave-one-subject-out validation across seven camera-view conditions (three single-view and four multi-view conditions) and four ROI strategies. Overall, root mean square error was ~4.7–5.6 N for the horizontal and mediolateral force components and ~10.6–11.0 N for the vertical component. Including the handled object as a second ROI generally improved force estimation, with some of the largest benefits under single-camera conditions, whereas pixel-level segmentation provided little additional improvement. Multi-camera capture provided the clearest benefit for peak-force estimation, particularly for the vertical component, whereas differences in overall frame-level error among camera configurations were comparatively modest. These findings demonstrate the feasibility of estimating continuous, bilateral, directional hand-force estimates from RGB video and known load mass without requiring sensors on the worker or handled objects as model inputs, supporting the development of more scalable occupational physical exposure and risk assessments.

## 1.0 Introduction

Work-related musculoskeletal disorders (WMSDs) remain a major occupational health concern in labor-intensive industries and continue to account for substantial workplace injuries and lost workdays (Lan et al., 2026; Liberty Mutual Insurance, 2023; U.S. Bureau of Labor Statistics, 2024). Workers engaged in manual material handling (MMH) tasks, including lifting, carrying, pushing, and pulling, are at elevated risk of developing WMSDs (Blasco-Abadía et al., 2025; Da Costa & Vieira, 2010; Yang et al., 2016). Because MMH tasks remain common across sectors such as manufacturing, construction, warehousing, and logistics, effective assessment of physical exposures is vital (Da Costa & Vieira, 2010). Physical exposure assessment can help identify high-risk tasks, support the development of ergonomic interventions, and improve understanding of exposure–risk relationships (Garosi et al., 2025; Javanmardi et al., 2025; Mathiassen & Winkel, 1991; Spielholz et al., 2001; Wells et al., 1997; Winkel & Mathiassen, 1994). While these assessments often focus on postural demands, external hand forces—the loads applied at the hands during interactions with objects—have been identified as a critical determinant of biomechanical loading and WMSD risk during MMH tasks (Koppelaar & Wells, 2005; Wells et al., 2004).

Although several approaches exist for assessing occupational exposures, estimating external hand forces during MMH tasks remains particularly challenging. Conventional approaches include self-reports, observational methods, and direct measurements (Burdorf & van der Beek, 1999; Koppelaar & Wells, 2005; Wells et al., 2004). Self-report and observational methods can be used to evaluate physical workload through questionnaires, checklists, and rating scales; however, these approaches typically provide only coarse or subjective estimates of physical exposures (Kumar, 1993; Spielholz et al., 2001; Wiktorin et al., 1996). Such methods may thus be useful for preliminary screening or broad ergonomic assessments, but they often lack the precision required for detailed biomechanical analysis (Kumar, 1993; Wiktorin et al., 1993). Direct measurements of hand forces are also possible using instrumentation such as load cells, force gauges, or instrumented objects (Burdorf, 1995; Van Der Beek & Frings-Dresen, 1998). However, implementing such measurements in real workplace environments is often difficult because they require specialized equipment, careful calibration, and controlled experimental setups that may interfere with natural worker behavior (Koppelaar & Wells, 2005; Wang et al., 2023). Consequently, obtaining accurate, continuous hand force measurements in real-world MMH tasks remains a persistent challenge.

Recent technological developments have enabled alternative approaches for estimating hand forces during MMH tasks, often relying on wearable sensors, markerless motion capture systems, and machine learning methods (Hlucny & Novak, 2020; Lee et al., 2020; Lim & D'Souza, 2019; Ojelade, 2024). Inertial measurement units, for instance, have been used to classify load levels during load carriage and related handling activities (H. Lee et al., 2020; Lim, 2024; Lim & D'Souza, 2019), while wearable electromyography systems have been used to estimate hand forces or muscle loads during manual tasks (Mobasser & Hashtrudi-Zaad, 2012; Saponas et al., 2008; Taori & Lim, 2024). Markerless motion capture systems have also been used to derive body kinematics from video for estimating external hand forces and other physical exposure metrics during MMH tasks (Ojelade, 2024; Steinebach et al., 2020; Zhu et al., 2025). However, wearable approaches can introduce user burden, discomfort, calibration and battery-management demands, and added cost (Dempsey et al., 2005; Smith et al., 2024; Zhuang et al., 2019), while markerless motion capture systems often still rely on specialized equipment and controlled settings (Kubota et al., 2026; Wade et al., 2022). In addition, these markerless approaches primarily emphasize body kinematics and skeletal representations and do not explicitly model human–object interactions, even though such interactions are important for assessing physical exposure during MMH tasks (Bezzini et al., 2023; Paudel et al., 2022; W. Zhao et al., 2023; Zheng et al., 2021). These limitations reduce the practicality of such approaches for large-scale or real-world workplace applications.

Specific advances in computer vision and deep learning have created new opportunities for estimating physical exposures, including external hand forces, directly from video without wearable sensors or instrumented objects. In fact, visual information about hand–object motion and interaction can be used to infer contact forces during hand–object interaction (Pham et al., 2015, 2018), and hand pressure patterns can be estimated from RGB images using visual cues and hand representations (Tang et al., 2025; Y. Zhao et al., 2025). Together, these results suggest that vision-based methods could provide a useful alternative for estimating aspects of physical workload without requiring direct force instrumentation.

Since hand force magnitude and direction are reflected in the relative motion and positions of the worker and handled object (Louis et al., 2022; Pham et al., 2015, 2018), *vision–language models* (VLMs)-based approaches could support external hand force estimation by identifying task-relevant visual information associated with these interactions. VLMs integrate visual representations with language-based contextual understanding, enabling text-guided localization of the body segments and objects involved in force-generating contact (e.g., hands, box; Kang et al., 2025; Radford et al., 2021). Given the value of detailed force information for physical exposure assessment and biomechanical modeling (Faber et al., 2013, 2018), it is important to determine how accurately such VLM-based approaches can estimate dynamic, triaxial, bilateral hand forces. Estimating such forces is particularly challenging because they vary continuously over time, act along three orthogonal axes, and must be estimated independently for each hand (Faber et al., 2013, 2018). It is also important to determine how the selection of regions of interest (ROIs) and camera viewpoint affect force estimation performance. ROI and object representations determine which visual features are emphasized during inference (Kirillov et al., 2023; Liu et al., 2024), whereas camera viewpoint affects the visual perception of motion, spatial relationships, and object interactions (Rajabi et al., 2026; Sigal et al., 2010; Tang et al., 2025). Therefore, our primary objective in the current study was to evaluate the accuracy of a VLM-based approach for estimating *dynamic, triaxial, bilateral external hand forces* during diverse MMH tasks. As a secondary objective, we investigated how different *ROI Strategies* and *Camera View Conditions* affected force estimation performance. Our results were intended to support the future development of practical, VLM-based methods for estimating physical exposures in occupational environments.

## 2.0 Methods

### 2.1 Overview of the Dataset

We used a subset of data obtained from a prior study by Ojelade et al. (2025), in which 35 healthy young adults (21 males and 14 females) performed eight simulated MMH tasks under various conditions in a controlled laboratory setting. See Appendix A.1 for participant demographics and inclusion criteria. We focused here on five MMH tasks to represent some common occupational activities: 1) symmetric box lifting from the floor or knee height to individual hip height (Task 1); 2) asymmetric box lifting from a table placed in front of the participant to a table positioned at a 90° angle to the left (Task 2); 3) box carrying over a distance of 2.4 m (Task 3); 4) box pushing over a distance of approximately 0.7 m on a table set at 0.74 m height (Task 4); and 5) box pulling toward the body over the same distance and table height (Task 5).

Participants completed these tasks in multiple task conditions involving two *Hand Configurations* (broad = 52 cm handle spacing and narrow = 33 cm handle spacing; see Figure A.1), three *Box Mass* levels (6, 9, and 12 kg), and two *Lift Origins* (floor and knee height). Each condition was performed twice (2 *Lift Origins* × 2 *Hand Configurations* × 3 *Box Masses* × 2 replications = 24 trials per participant). A single wooden box (width = 26.0 cm; depth = 41.0 cm; height = 23.5 cm), instrumented with bilateral triaxial load cells (Michigan Scientific Corp. TR3D-A-1K, Charlevoix, MI, USA), was used during all tasks (see Appendix A.2 for MMH task details and experimental procedures). Whole-body kinematics were recorded using three synchronized Azure Kinect™ cameras (Microsoft Corporation, Seattle, WA, USA) positioned approximately 1.74 m from the edge of the work area. The cameras recorded RGB video streams at 30 Hz (1280 × 720 pixels) and provided three distinct viewpoints of the task area. Details regarding camera placement, synchronization, and instrumentation are provided in Appendix A.3. Time series of external hand force data—including bilateral anteroposterior ($F_x$), mediolateral ($F_y$), and vertical ($F_z$) components from each load cell (see Figure A.2 for coordinate system definition)—were sampled at 200 Hz. Raw data were acquired using a custom LabVIEW (National Instruments, Austin, TX, USA), processed to reduce sensor noise and artifacts, and temporally aligned with the RGB video recordings (see Appendix A.4). These dynamic force measurements served as the ground-truth target variables for our model development and evaluation.

### 2.2 Overview of the VLM-Based Pipeline for External Hand Force Estimation

We developed and evaluated multi-stage VLM-based pipelines to estimate bilateral external hand forces from the RGB video streams (Figure 1). The overall pipeline was inspired by the VLM-based approach developed in Rajabi et al. (2026), which was designed to estimate horizontal and vertical hand distances during symmetric box lifting task, but was adapted here for continuous external hand force estimation. The pipeline consisted of three primary steps. Step 1 involved detecting and segmenting task-relevant ROIs, Step 2 involved extracting visual features from the identified ROIs, and Step 3 involved estimating bilateral hand forces using a transformer-based regression model. Each step is described in detail in the following sections. All processing was performed offline using Python (v3.10.11; https://www.python.org) on secure Virginia Tech Advanced Research Computing resources (VT ARC; https://arc.vt.edu/); no participant videos, images, or derived data were uploaded to or processed on external or cloud-based systems. RGB video handling and frame extraction were implemented using OpenCV (https://opencv.org/), and model inference and training were conducted using PyTorch (https://pytorch.org/; Paszke et al., 2019).

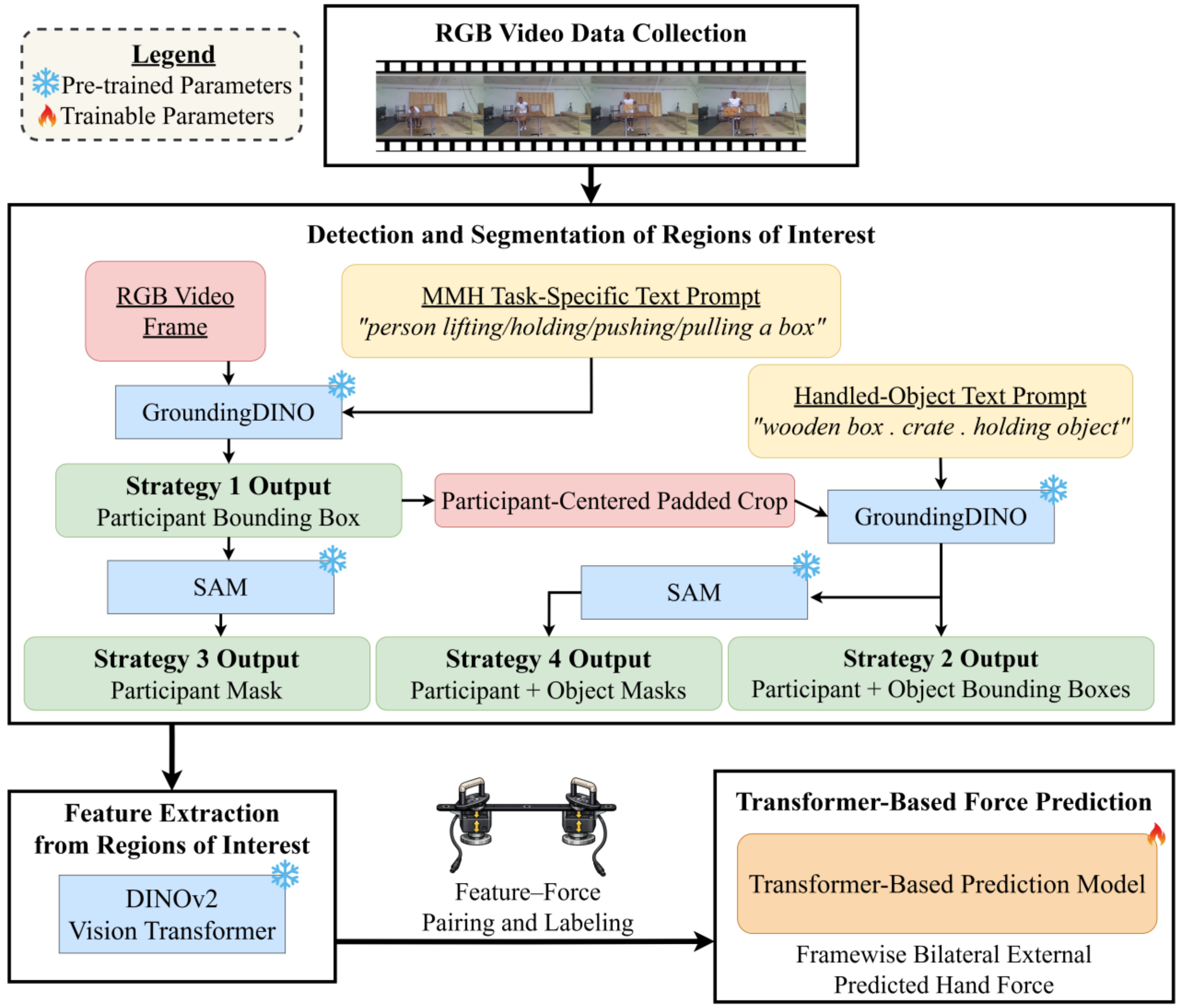

Figure 1. Overview of the pipeline for estimating bilateral external hand forces from RGB video during MMH tasks. Task-relevant regions of interest (ROIs) were localized using a text-guided detection process with GroundingDINO and, where applicable, refined using bounding-box–guided segmentation with SAM. ROI features extracted using DINOv2 were then paired with synchronized load-cell force measurements and processed using a transformer-based prediction model to estimate frame-level bilateral external hand forces.

### 2.3 Detection and Segmentation of Regions of Interest (ROIs)

To isolate MMH task-relevant visual information from each RGB frame, ROIs corresponding to the participant and handled object were detected using GroundingDINO (Liu et al., 2024), a zero-shot, open-set object detector. To assess the effect of ROI representation on force estimation performance, four alternative ROI processing strategies were evaluated. Across all strategies, the participant was first detected in each frame using an MMH task-specific textual query. Specifically, the prompt *"person lifting a box"* was used for Tasks 1 and 2, *"person holding a box"* for Task 3, *"person pushing a box"* for Task 4, and *"person pulling a box"* for Task 5. When multiple detections were returned, the bounding box with the highest confidence score was retained as the participant region. For strategies requiring handled-object localization, a padded crop, centered on the detected participant region, was then generated to suppress background interference and to constrain the search space for the handled object. GroundingDINO was subsequently applied within this cropped region using the prompt *“wooden box . crate . holding object”* to detect the handled object. The resulting bounding boxes defined candidate ROIs corresponding to the participant and, when applicable, the handled object.

In Strategy 1 (S1), only the participant bounding box detected by GroundingDINO was retained as the ROI. In Strategy 2 (S2), both the participant and handled-object bounding boxes were retained as ROIs. Strategies 3 and 4 (S3 and S4) extended S1 and S2, respectively, by refining the detected bounding boxes using the Segment Anything Model (SAM; ViT-H backbone; Kirillov et al., 2023), a promptable segmentation model that generates pixel-level masks from bounding-box prompts. In S3, SAM was applied to the participant bounding box to generate a pixel-level participant mask, whereas SAM was applied in S4 to both the participant and handled-object bounding boxes to generate pixel-level masks for both regions. Figure 2 shows frames from View 1 (V1), illustrating the five MMH tasks and the four ROI representation strategies; corresponding examples from Views 2 (V2) and 3 (V3) are provided in Figures A.3 and A.4, respectively.

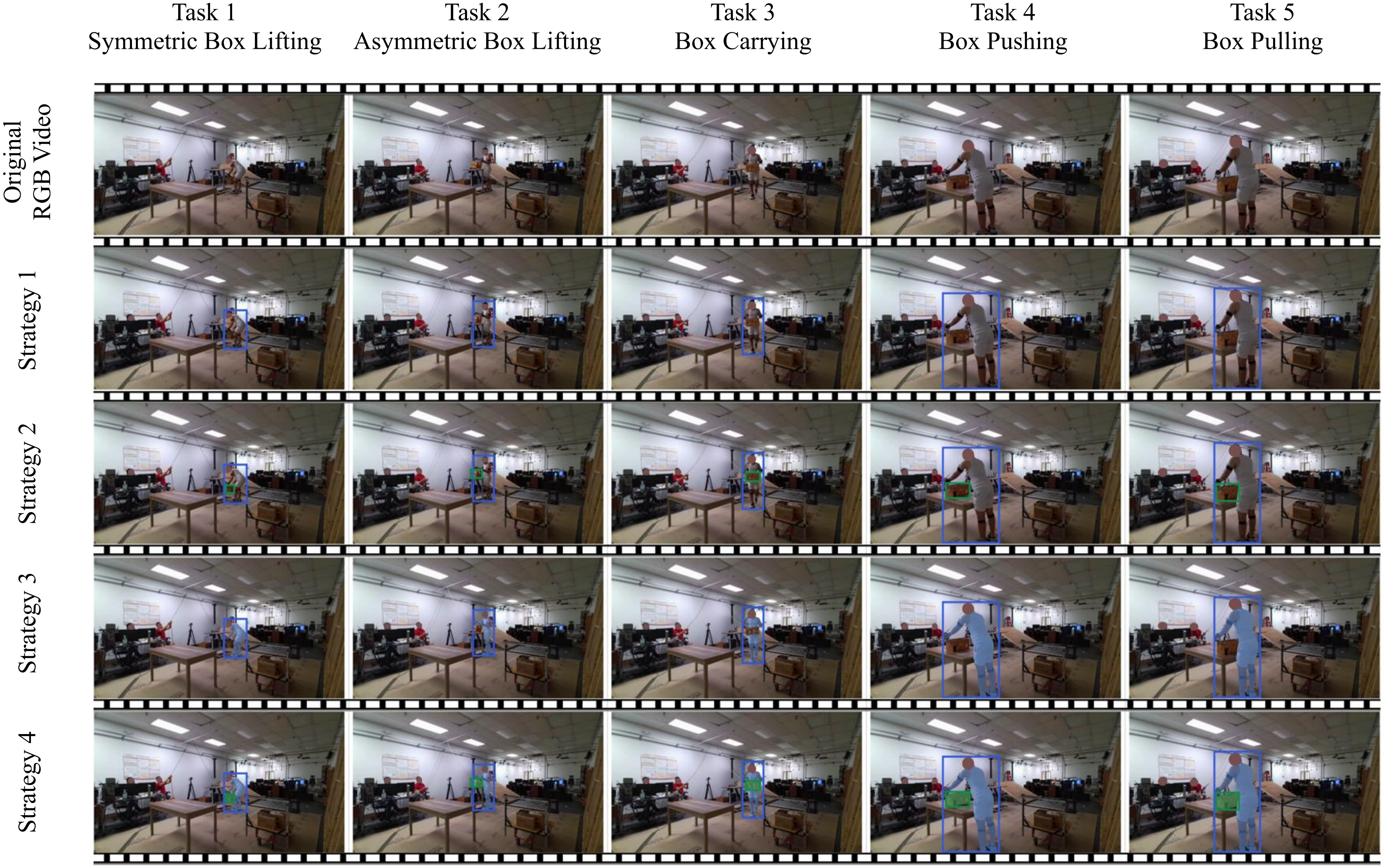

Figure 2. Example frames from View 1 (V1) illustrating the five manual material handling (MMH) tasks and the four region-of-interest (ROI) representation strategies evaluated in this study. The first row shows the original RGB frames extracted from the video recordings. Subsequent rows illustrate the four ROI processing strategies: Strategy 1, participant bounding box; Strategy 2, participant and handled-object bounding boxes; Strategy 3, participant segmentation mask; and Strategy 4, participant and handled-object segmentation masks. Participant regions are shown in blue and handled-object regions in green.

### 2.4 Feature Extraction from ROIs

Visual features were extracted from each detected ROI using DINOv2 (ViT-Base; Oquab et al., 2024), a self-supervised vision transformer producing general-purpose visual representations without requiring labeled training data. For each frame, detected ROIs were cropped from the original RGB frame based on bounding box coordinates or, in the case of segmentation-based strategies (S3 and S4), based on the bounding rectangle of the pixel-level mask. Prior to feature extraction, each cropped region was resized to 224 × 224 pixels to match the input resolution used for DINOv2 feature extraction, and then normalized using ImageNet channel-wise statistics (Deng et al., 2009). DINOv2 was subsequently applied in a zero-shot configuration with frozen weights, yielding a 768-dimensional feature vector per ROI. For strategies involving a single ROI (S1, S3), features were stored as a tensor of shape $[T, 1, 768]$, where $T$ is the number of frames in the video sequence for a given trial. For strategies involving two ROIs (S2, S4), features from the participant and handled-object regions were stacked to produce a tensor of shape $[T, 2, 768]$; frames in which the handled object was not detected were zero-padded to maintain a uniform tensor shape.

### 2.5 Feature–Force Pairing and Labeling

Visual features extracted from the detected and, where applicable, segmented ROIs in each RGB frame were paired with corresponding load-cell measurements to generate synchronized feature–force pairs for model training. The start and end frames defining each MMH task segment were manually identified by two research assistants. Six bilateral load-cell channels were used as regression targets: $F_x$, $F_y$, and $F_z$ for each hand. The resulting datasets therefore consisted of temporally aligned sequences of ROI-based visual features and six channels of corresponding frame-level bilateral force labels.

### 2.6 Transformer-Based Force Prediction

Transformer-based force prediction models were trained to estimate bilateral external hand forces from the temporally ordered visual features extracted from RGB video frames. The force prediction process involved designing the transformer-based model architecture, defining data handling procedures, training and validation strategies, and evaluating model performance, with details as follows.

#### 2.6.1 Model Architecture

We used a modified transformer encoder model (Vaswani et al., 2017) to estimate bilateral external hand forces while explicitly modeling temporal dependencies across sequences of consecutive video frames. Considering sequences of frames—rather than treating each frame independently—is important because external hand forces during MMH tasks evolve dynamically, and accurate frame-level estimation often depends on surrounding temporal context that captures the dynamics of body motion and object interactions over time (Carreira & Zisserman, 2017).

The model consisted of six components applied sequentially. First, an ROI fusion layer merged the framewise DINOv2 feature vectors into a single frame-level feature embedding. For single-ROI strategies (S1, S3), the 768-dimensional feature vector was projected directly through a linear layer; for dual-ROI strategies (S2, S4), the participant and handled-object feature vectors were concatenated to form a 1536-dimensional input prior to linear projection. In both cases, the projection—followed by layer normalization, Gaussian error linear unit (GELU) activation (Hendrycks & Gimpel, 2016) and dropout—produced a 256-dimensional feature embedding per frame. Second, the normalized load mass (in kg) was embedded through a learned linear projection and broadcast-added to all frame tokens, providing the model with task load context at every time step. Third, a learnable camera-view feature embedding—one per view (V1, V2, V3)—was similarly broadcast-added to encode the spatial perspective of the camera from which each sequence was recorded; view-specific sequences were processed independently without cross-view feature fusion. Fourth, sinusoidal positional encodings (Vaswani et al., 2017) were added to the feature embedding sequence to preserve temporal ordering across frames. Fifth, the resulting sequence was processed by a stack of four transformer encoder layers, each comprising eight attention heads, a feedforward sublayer of dimension 512, GELU activation (Hendrycks & Gimpel, 2016), pre-layer normalization, and dropout regularization. An attention mask was applied during encoding so that padded frames did not contribute to the self-attention computation. Sixth, a per-frame regression head—consisting of two fully connected layers with a GELU activation and dropout—mapped each encoder output to simultaneous estimates of the six bilateral force channels. Further details of the model architecture are provided in Appendix A.5.

### 2.6.2 Data Handling, Model Training, and Validation

Each input sequence was set to a fixed length of 300 consecutive frames, selected as a compromise to balance computational efficiency and MMH task coverage. Preprocessing steps including frame zero-padding and attention masking were applied (see Appendix A.5). Force labels were normalized prior to training on a per-channel basis; load mass values were similarly standardized. Normalization statistics were computed exclusively from training data and were applied consistently to the validation and test partitions (described below).

Model training was performed for up to 100 epochs, with early stopping applied to prevent overfitting (Prechelt, 1998). Specifically, training was terminated if the validation loss—the mean squared error computed on a held-out validation data—failed to improve over 20 consecutive epochs. In this case, the model state corresponding to the minimum validation loss was retained and used for evaluation on the held-out test participant. A leave-one-subject-out (LOSO) cross-validation strategy was used for model validation; training was replicated across 35 folds, each fold using data from 34 participants for training and the remaining participant for testing (e.g., Ojelade et al., 2025; Porta et al., 2021; Rajabi et al., 2026, 2027). Within each fold, 10% of training participants were further withheld as a within-fold validation set used exclusively for early stopping and checkpoint selection. Although LOSO can introduce higher variance due to individual differences in task execution, it more closely replicates real deployment scenarios in which models trained on known participants are applied to an unseen individual (Gholamiangonabadi et al., 2020; Jordao et al., 2018).

### 2.6.3 Evaluation Metrics

Force estimation performance was evaluated separately for each of the six bilateral force channels. Three metrics were computed (see Appendix A.5 for metric definitions): mean absolute error (MAE), root mean square error (RMSE), and Peak Error (PE). MAE and RMSE quantify frame-level estimation accuracy, with RMSE applying greater weight to larger individual errors. PE was defined as the difference between the 95$^{th}$ percentile of the absolute actual forces and the 95$^{th}$ percentile of the absolute predicted forces; it was included to capture the ability of the model to reproduce peak force magnitudes that are most relevant for ergonomic risk assessment.

### 2.7 Statistical Analyses

Separate linear mixed-effects models (LMMs) were fitted for each evaluation metric and force channel to examine the effects of *MMH Task*, *Hand Configuration*, *Box Mass*, *Camera View Condition*, *ROI Representation Strategy*, and *Biological Sex* (as a blocking factor). Analyses were limited to three-way interactions, since higher-order interactions are generally difficult to interpret meaningfully. To meet parametric model assumptions, MAE and RMSE values were log-transformed to achieve normal residuals. MAE and RMSE were strongly correlated across all force channels; for efficiency, only RMSE results are reported below, with complete MAE results provided in Appendix Tables A.1–A.3. All analyses were performed in JMP Pro 19 (SAS Institute Inc., Cary, NC) with statistical significance determined at $p < 0.05$. Significant main and interaction effects were followed by *post hoc* pairwise comparisons using Tukey's HSD test.

## 3.0 Results

Statistical results for all main and interaction effects are summarized in Tables A.1–A.3, with corresponding figures in Figures 4–9 and A.5–A.13. An example of continuous actual and predicted bilateral external hand forces is shown in Figure 3 for all three force directions and four *ROI Strategies*.

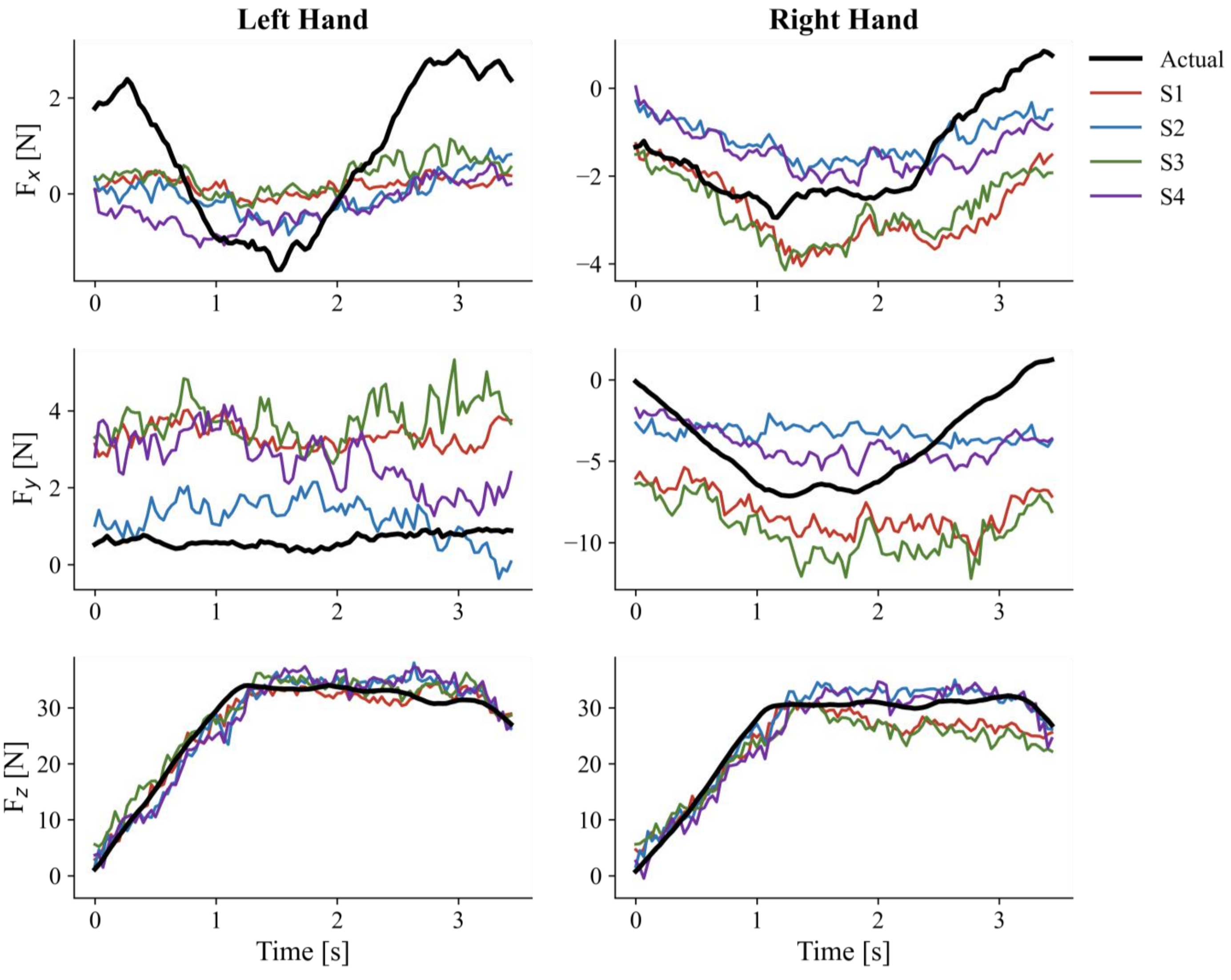

Figure 3. Example of continuous bilateral external hand force predictions during Task 3 (box carrying) with the narrow *Hand Configuration,* 6 kg *Mass*, and View 1 (V1), across the four *ROI Strategies* (S1–S4). Actual and predicted forces are shown for the horizontal ($F_x$; top), mediolateral ($F_y$; middle), and vertical ($F_z$; bottom) directions.

Because our primary methodological interest was on the influences of *Camera View Condition* and *ROI Strategy*, the results presented below emphasize main or interaction effects involving at least one of these factors, evaluated using RMSE and PE (summarized in Table 1). Effects of *MMH Task* and *Mass* are noted briefly, as context for interpreting force estimation performance.

Table 1. Summary *Camera View Condition* and *ROI Strategy* effects on force estimation performance across force axes and hands. ✓ indicates a statistically significant effect.

| Outcome Measure | | | **Camera View (Main Effect)** | **Camera View × MMH Task** | **ROI Strategy (Main Effect)** | **ROI Strategy × MMH Task** | **Camera View × ROI Strategy** |
|---|---|---|---|---|---|---|---|
| $F_x$ | RMSE | Right | ✓ | ✓ | ✓ | ✓ | |
| | | Left | ✓ | ✓ | ✓ | ✓ | |
| | PE | Right | ✓ | ✓ | ✓ | | |
| | | Left | ✓ | ✓ | ✓ | ✓ | ✓ |
| $F_y$ | RMSE | Right | ✓ | | ✓ | | |
| | | Left | | | | | |
| | PE | Right | ✓ | | | ✓ | |
| | | Left | ✓ | | | ✓ | |
| $F_z$ | RMSE | Right | ✓ | ✓ | ✓ | ✓ | |
| | | Left | ✓ | ✓ | ✓ | ✓ | |
| | PE | Right | ✓ | ✓ | ✓ | | ✓ |
| | | Left | ✓ | ✓ | ✓ | | |

### 3.1 Effects of *Mass* and *MMH Task* on Force Estimation Performance

There was a significant main effect of *Mass* on RMSE and PE for all force directions and both hands (Tables A.1–A.3, Figure A.5), with errors increasing consistently from 6 to 9 to 12 kg. RMSE was roughly 26–36% smaller at 6 kg vs. 12 kg across all force channels and both hands. PE was also smaller at 6 kg vs. 12 kg, with differences of ~21–49% for $F_x$ and $F_y$ and ~83–87% for $F_z$. There was also a significant main effect of *MMH Task* on RMSE and PE for all force directions and both hands (Figure A.6), with smaller $F_x$ RMSE and PE in Tasks 1, 2, and 3 vs. Tasks 4 and 5; larger $F_y$ RMSE and PE in Tasks 2 and 3 vs. Tasks 1, 4, and 5; and smaller $F_z$ RMSE in Tasks 1 and 3 vs. Tasks 2, 4, and 5; $F_z$ PE was smallest in Task 3, but was significantly larger in Task 1 vs. in Task 5.

### 3.2 $F_x$: Effects of *Camera View Condition* and *ROI Strategy*

**RMSE**

Table A.1 summarizes LMM outcomes for $F_x$. There was a significant *Camera View Condition* × *MMH Task* interaction effect on $F_x$ RMSE for both hands (Figure 4). The magnitude of RMSE differences among *Camera View Conditions* was generally small within Tasks 1–4, and somewhat larger (~10–13%) in Task 5, with no *Camera View Condition* showing a consistent advantage.

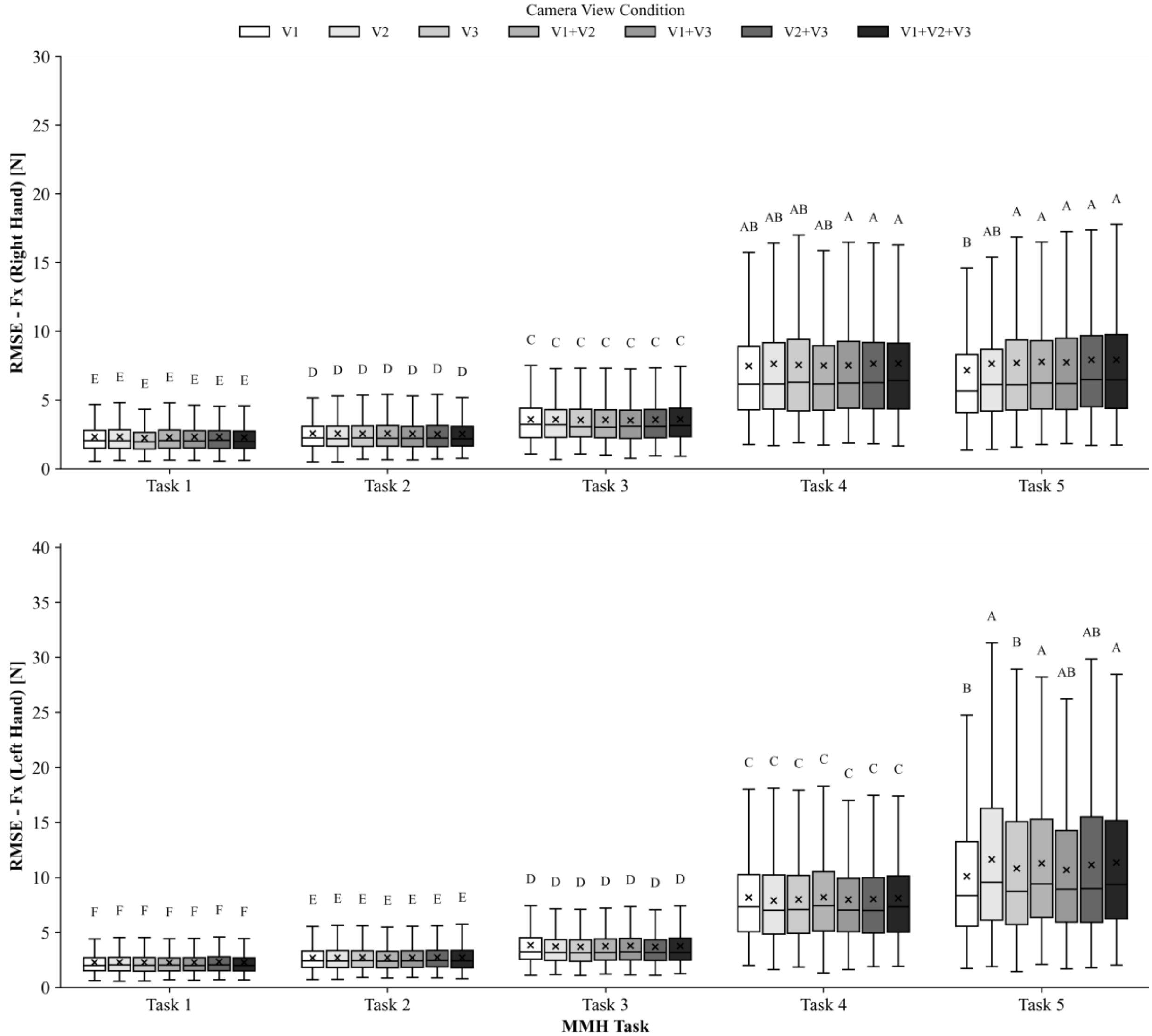

Figure 4. *Camera View Condition* × *MMH Task* interaction effect on $F_x$ RMSE for the right (top) and left (bottom) hands. Based on *post hoc* pairwise comparisons, boxes that do not share a common letter are significantly different.

There was also a significant *ROI Strategy* × *MMH Task* interaction effect on $F_x$ RMSE for both hands (Figure 5). The dual-ROI strategies (S2, S4) generally produced smaller RMSE vs. the single-ROI strategies (S1, S3) in Tasks 4 and 5, though differences among strategies were small in Tasks 1–3.

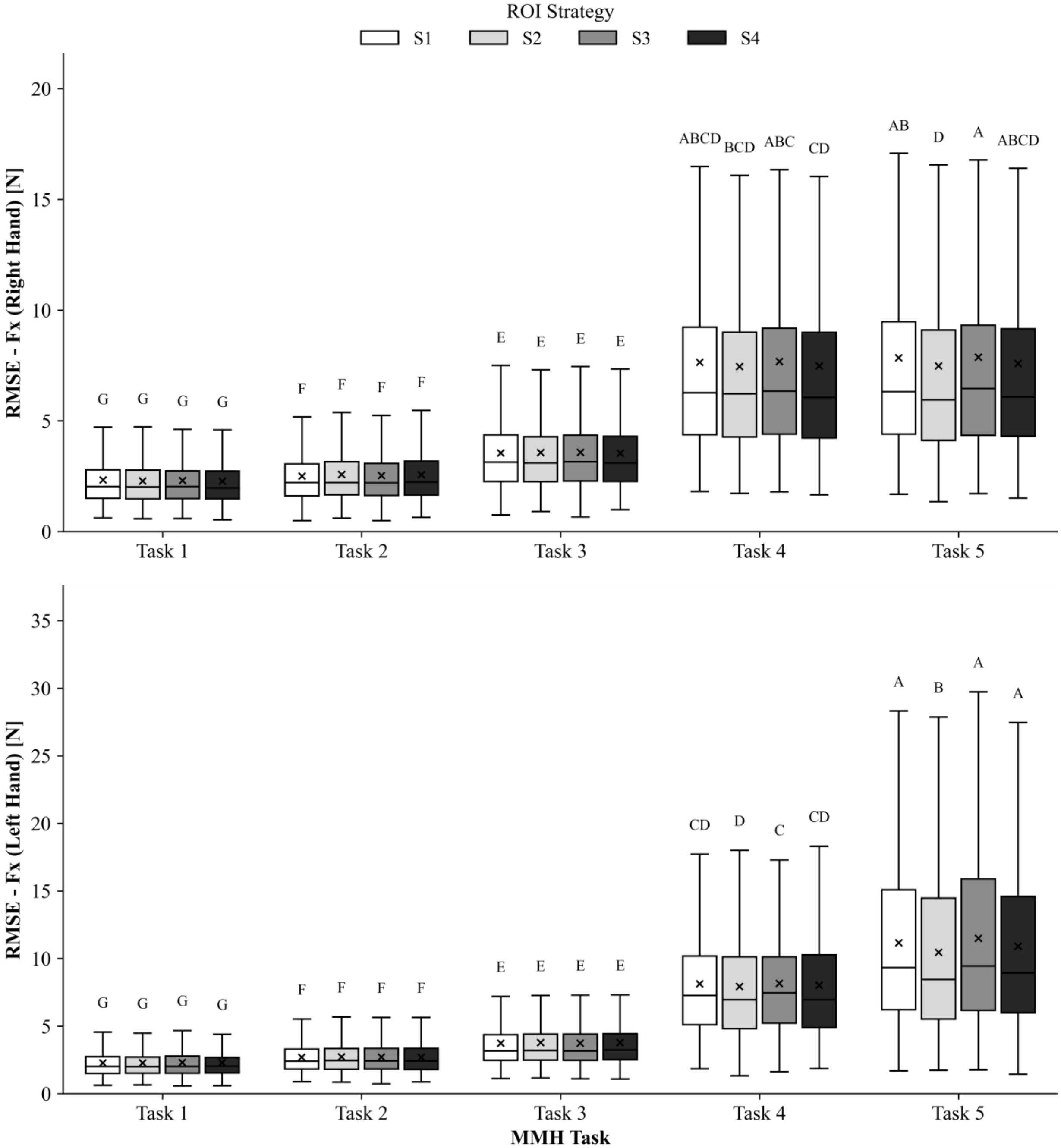

Figure 5. *ROI Strategy* × *MMH Task* interaction effect on $F_x$ RMSE for the right (top) and left (bottom) hands. Based on *post hoc* pairwise comparisons, boxes that do not share a common letter are significantly different.

## PE

There was a significant *Camera View Condition* × *MMH Task* interaction effect on $F_x$ PE for both hands (Figure 6). PE values were typically positive across *Camera View Condition* and *MMH tasks*, with generally small differences among *Camera View Conditions* within Tasks 1–3 and larger differences in Tasks 4 and 5, though no *Camera View Condition* consistently produced smaller PE across *MMH tasks*.

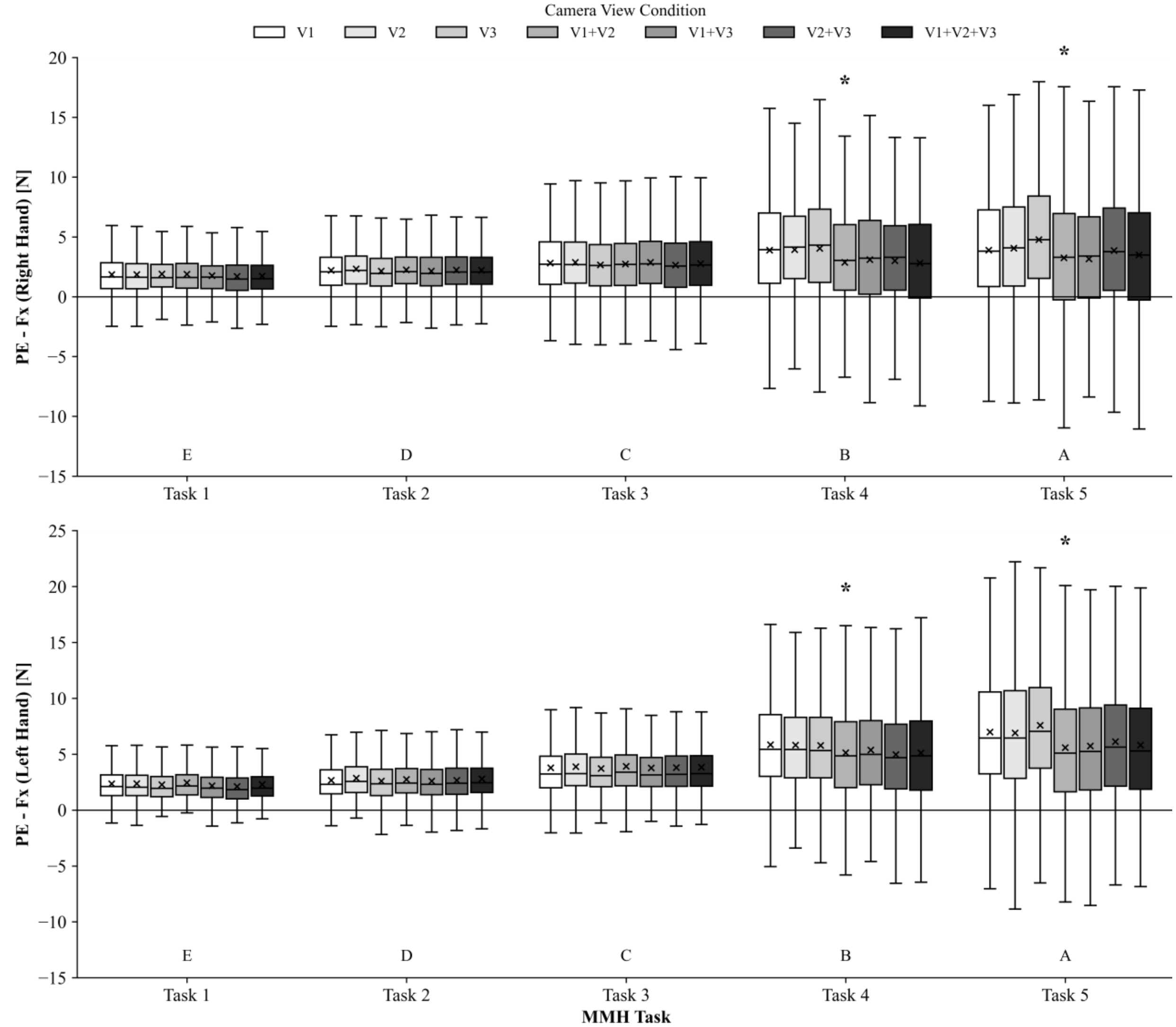

Figure 6. *Camera View Condition* × *MMH Task* interaction effect on $F_x$ PE for the right (top) and left (bottom) hands. Based on *post hoc* pairwise comparisons, *MMH Tasks* that do not share a common letter are significantly different. Asterisks denote a significant effect of *Camera View Condition* within a specific task.

For the right hand, there was a significant main effect of *ROI Strategy* on $F_x$ PE, with S2 and S4 showing smaller PE vs. S1 and S3 (S1: ~2.9 N, S3: ~2.9 N vs. S2: ~2.7 N, S4: ~2.7 N). For the left hand, there was a significant *ROI Strategy* × *MMH Task* interaction effect on $F_x$ PE (Figure A.7); the pattern was similar to RMSE, with S2 and S4 showing smaller PE vs. S1 and S3, most notably in Tasks 4 and 5. There was also a significant *ROI Strategy* × *Camera View Condition* interaction effect on PE for the left hand (Figure A.8), though PE differences among ROI strategies remained small within each *Camera View Condition*.

### 3.3 $F_y$: Effects of *Camera View Condition* and *ROI Strategy*

Table A.2 summarizes LMM outcomes for $F_y$. There was a significant main effect of *Camera View Condition* on $F_y$ RMSE for the right hand (Figure A.9; ~4.6–4.8 N across all *Camera View Conditions*); this main effect was not significant for the left hand (~5.6–5.7 N across all *Camera View Conditions*). There was a significant main effect of *Camera View Condition* on $F_y$ PE for both hands (Figure A.10), with PE values generally smaller under multi-camera conditions vs. single-camera conditions (right hand: ~2.3–2.5 N vs. ~2.7 N; left hand: ~3.1–3.3 N vs. ~3.3–3.5 N). There was a significant main effect of *ROI Strategy* on $F_y$ RMSE for the right hand (Figure A.11; S1: ~4.7 N, S2: ~4.7 N, S3: ~4.8 N, S4: ~4.7 N), but no significant main effect of *ROI Strategy* on $F_y$ PE for either hand, with values of ~2.4–2.6 N for the right hand and ~3.2–3.3 N for the left hand across all *ROI Strategies*. There was also a significant *ROI Strategy* × *MMH Task* interaction effect on $F_y$ PE for both hands (Figure A.12), though *ROI Strategy* differences were small across *MMH tasks*.

### 3.4 $F_z$: Effects of *Camera View Condition* and *ROI Strategy*

**RMSE**

Table A.3 summarizes LMM outcomes for $F_z$. There was a significant *Camera View Condition* × *MMH Task* interaction effect on $F_z$ RMSE for both hands (Figure 7). Differences among *Camera View Conditions* were small (~3–11% across tasks), with no *Camera View Condition* showing a consistent advantage; differences tended to be larger in Tasks 2, and 3 and smallest in Tasks 4, and 5.

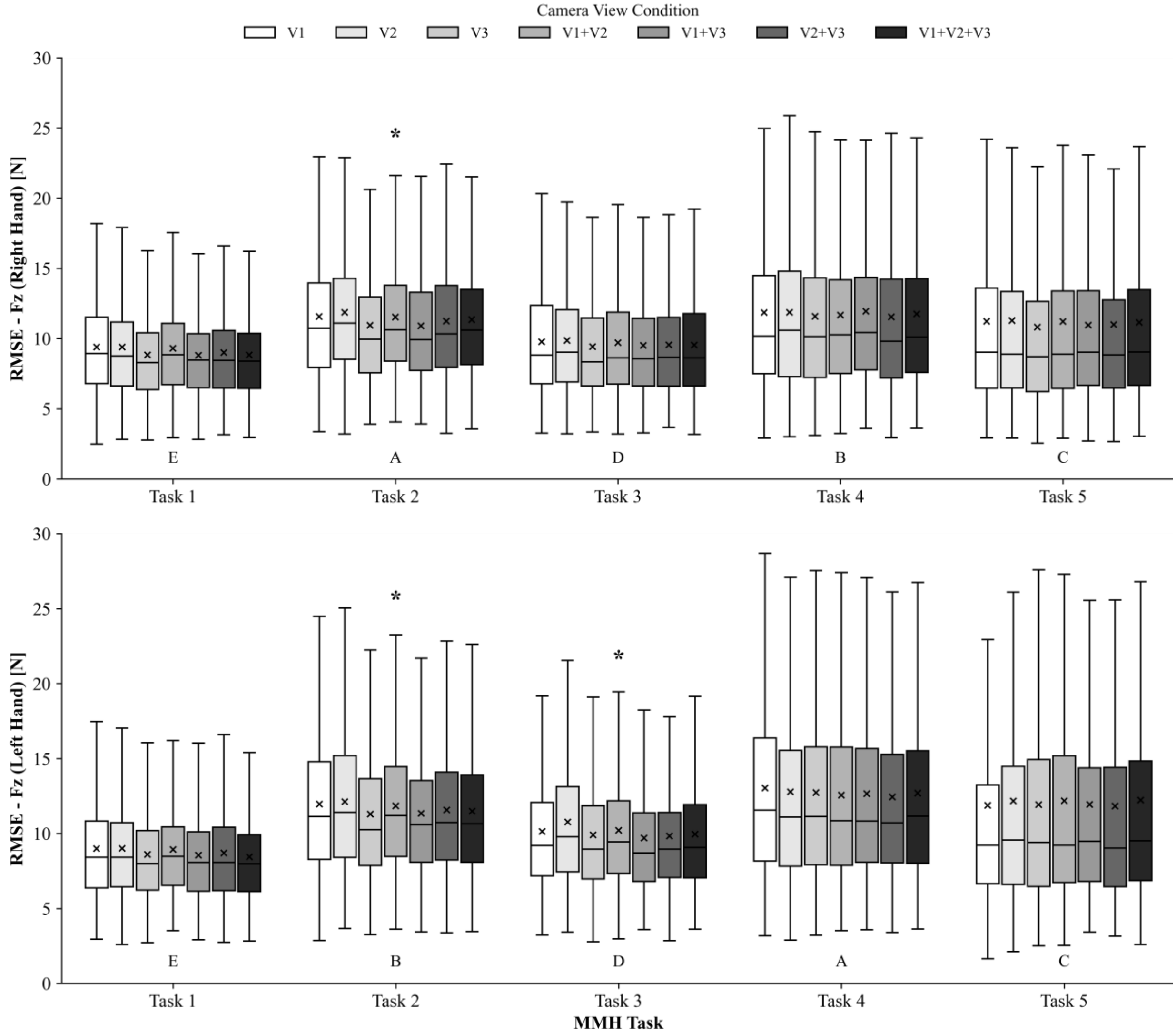

Figure 7. *Camera View Condition* × *MMH Task* interaction effect on $F_z$ RMSE for the right (top) and left (bottom) hands. Based on *post hoc* pairwise comparisons, *MMH Tasks* that do not share a common letter are significantly different. Asterisks denote a significant effect of *Camera View Condition* within a specific task.

There was also a significant *ROI Strategy* × *MMH Task* interaction effect on $F_z$ RMSE for both hands (Figure 8). The dual-ROI strategies (S2, S4) generally produced smaller RMSE vs. the single-ROI strategies (S1, S3), most notably in Tasks 1 and 2 (~5–7% lower), with smaller and less consistent differences in Tasks 3, and 5.

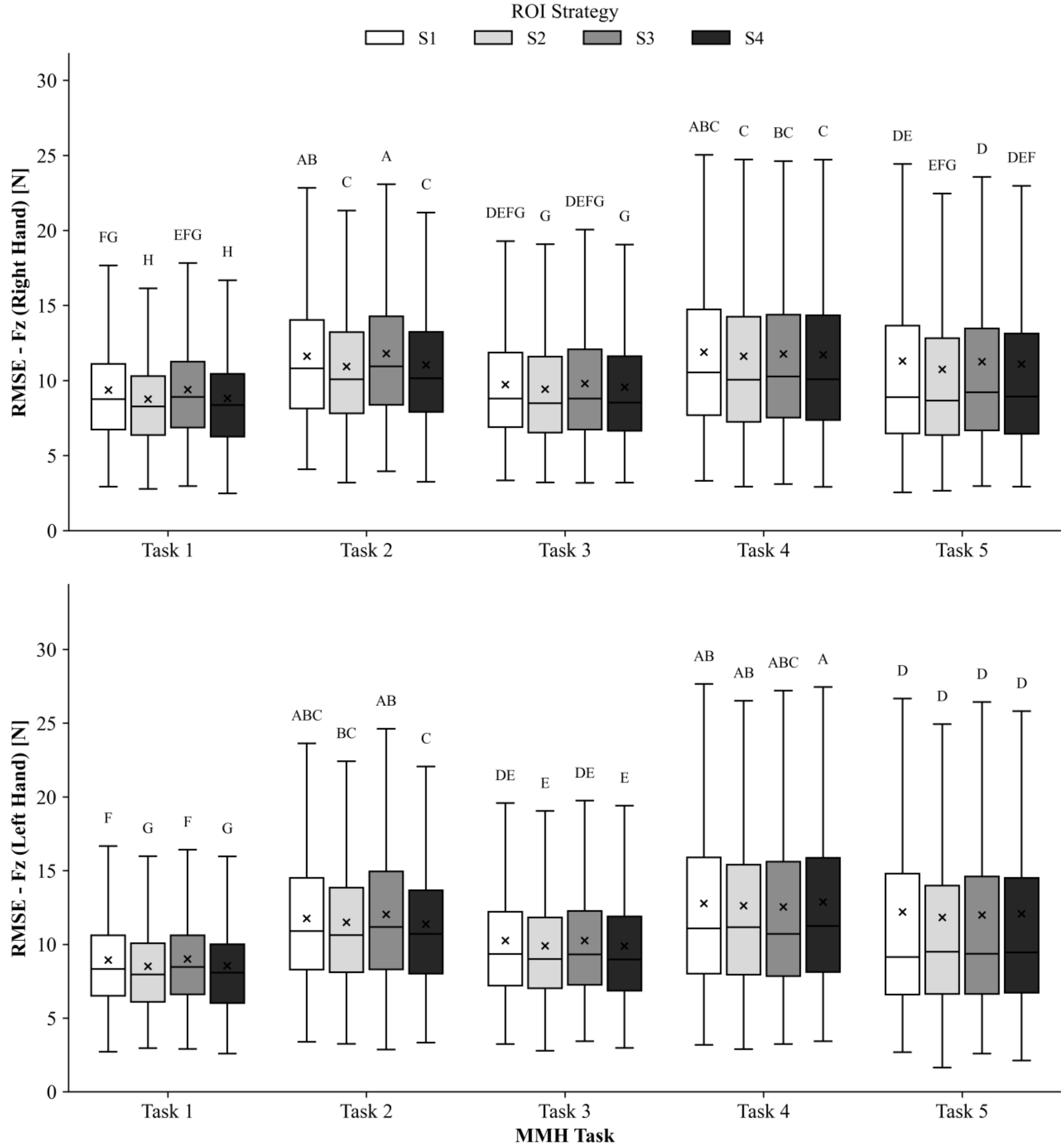

Figure 8. *ROI Strategy* × *MMH Task* interaction effect on $F_z$ RMSE for the right (top) and left (bottom) hands. Based on *post hoc* pairwise comparisons, boxes that do not share a common letter are significantly different.

## PE

There was a significant *Camera View Condition* × *MMH Task* interaction effect on $F_z$ PE for both hands (Figure 9). PE values were typically positive and were generally larger under single-camera conditions (V1, V2, V3) vs. multi-camera conditions, most pronounced in Tasks 2 and 4 (single-camera ~4.6–4.9 N vs. multi-camera ~1.9–2.3 N) and comparatively small in Task 3.

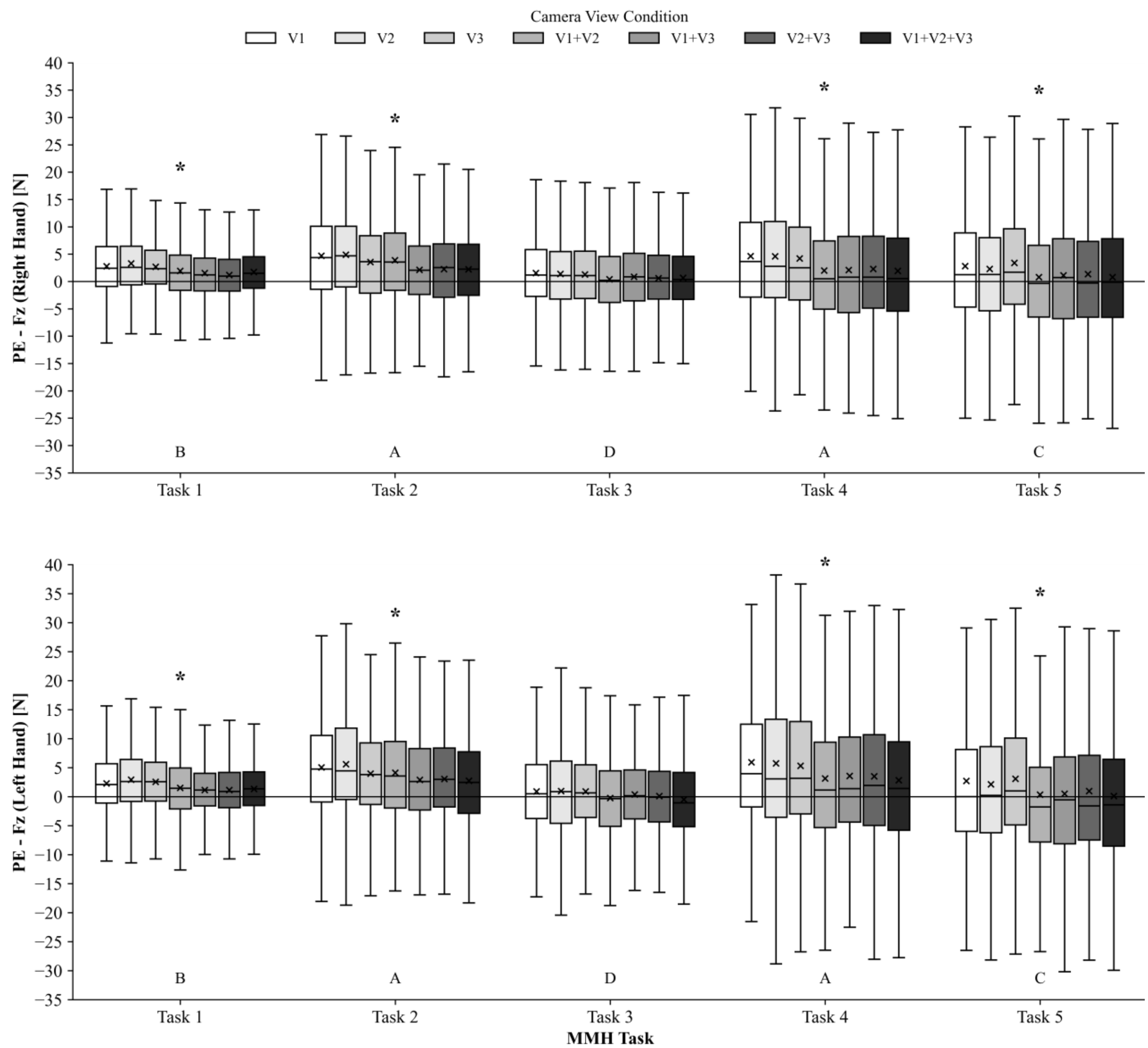

Figure 9. *Camera View Condition* × *MMH Task* interaction effect on $F_z$ PE for the right (top) and left (bottom) hands. Based on *post hoc* pairwise comparisons, *MMH Tasks* that do not share a common letter are significantly different. Asterisks denote a significant effect of *Camera View Condition* within a specific task.

For the right hand, there was also a significant *Camera View Condition* × *ROI Strategy* interaction effect on $F_z$ PE (Figure A.13); ROI Strategy differences were most evident under single-camera conditions, where S2 and S4 showed ~20–26% smaller PE vs. S1 and S3, while differences were small and inconsistent under multi-camera conditions.

## 4.0 Discussion

Our goal was to evaluate the feasibility of a VLM-based pipeline for estimating *dynamic, triaxial, bilateral* external hand forces from RGB video during diverse MMH tasks, and to investigate the effects of *Camera View Condition* and *ROI Strategy* on force estimation performance. We used an approach that integrated text-guided detection and segmentation, pretrained vision-transformer feature extraction, and transformer-based temporal regression, with performance evaluated using leave-one-subject-out validation. Overall, RMSE was ~4.7–5.6 N for $F_x$ and $F_y$ and ~10.6–11.0 N for $F_z$; the latter corresponded to ~21–32% of the static vertical load supported per hand across the 6–12 kg range examined. These findings support the feasibility of non-contact estimation of external hand forces from RGB video during forceful and dynamic MMH tasks. Below, we discuss these findings in relation to sensing modality, *MMH Task*, *Mass*, *Camera View Condition*, and *ROI Strategy*.

### 4.1 External Hand Force Estimation Using Different Sensing Modalities

Our per-hand RMSE values (~4.7–11 N, depending on the force axis) were broadly comparable to those reported using approaches requiring substantially more instrumentation. For example, Faber et al. (2013) reported RMSDs of ~11–27 N when estimating bimanual forces using force-plate and body-segment acceleration data. In other work, markerless motion capture combined with in-shoe pressures yielded RMSE of ~2.0–15.4 N horizontally, ~2.6–9.6 N laterally, and ~9.6–38.9 N vertically during the same five MMH tasks examined here (Ojelade, 2024). Although these comparisons are not direct, because the estimated force quantities and sensing configurations differed, our errors are within or, in some cases, below these ranges. Importantly, though, our methods required only RGB video and known box mass as model inputs, with no sensors needed on the worker or handled object. Further, our approach provided continuous force estimates independently for each hand and axis, rather than more crudely classifying discrete force or load levels. Together, these findings suggest that *detailed external hand-force estimates can be obtained with substantially less instrumentation than is typically required for direct or model-based force assessment* (Burdorf, 1995; Burdorf & van der Beek, 1999; Van Der Beek & Frings-Dresen, 1998; Wells et al., 2004).

Prior video-based work, though, has generally targeted related but different force quantities. Video-based kinetic approaches have estimated whole-body quantities such as ground reaction forces and joint moments (Kubota et al., 2026; Uhlrich et al., 2023), while other vision-based studies have estimated external or hand–object interaction forces or load levels (Louis et al., 2022; Pham et al., 2015, 2018; Zhu et al., 2025) and localized hand pressure from RGB images (Grady et al., 2022; Tang et al., 2025; Zhao et al., 2025). These approaches therefore differ from our present work, which estimates continuous bilateral and triaxial external hand forces during forceful, whole-body MMH tasks. This distinction is relevant because external hand forces are important inputs to biomechanical analyses of physical exposure during material handling (Koppelaar & Wells, 2005; Wells et al., 2004). Thus, our findings extend prior vision-based force estimation toward the dynamic, directional hand forces needed for detailed occupational exposure assessment.

### 4.2 Effects of *Mass* and *MMH Task*

RMSE and PE increased with *Mass* for all force channels. RMSE was ~26–36% smaller at 6 kg vs. 12 kg across all force channels and both hands, while PE was ~21–49% smaller for $F_x$ and $F_y$ and ~83–87% smaller for $F_z$ (Figure A.5). A similar increase in absolute hand-force estimation error with *Mass* was reported by Ojelade (2024) during these same MMH tasks using markerless motion capture and in-shoe pressure data. We suspect that increasing *Mass* altered lifting kinematics and posture (Lee, 2015), increasing variability in the participant–object motion represented by the visual features. Notably, this increase occurred even though box mass was provided explicitly as an input to our model.

*MMH Task* also affected all force channels, with smaller $F_x$ errors in Tasks 1–3 vs. Tasks 4 and 5, larger $F_y$ errors in Tasks 2 and 3 vs. Tasks 1, 4, and 5, and smaller $F_z$ errors in Tasks 1 and 3 vs. Tasks 2, 4, and 5 (Figure A.6). Estimation accuracy has previously been shown to vary by task phase and force direction (Faber et al., 2013). For $F_x$, the relatively larger errors during pushing and pulling were also reported by Ojelade (2024). That two structurally different approaches showed a similar task dependence suggests that the larger errors during pushing and pulling may reflect characteristics of these tasks rather than a limitation specific to either sensing approach.

### 4.3 Effects of *Camera View Condition*

*Camera View Condition* significantly affected nearly every force channel and metric, though the practical magnitude of these effects differed between metrics. The influences of view conditions on RMSE were modest, ~3–13% depending on task and force axis, and no single condition performed best across tasks (Figures 4 and 7; Figure A.9). Differences in PE were considerably larger, particularly for $F_z$, where PE was substantially smaller under multi-camera vs. single-camera conditions, most notably in Tasks 2 and 4 (~4.6–4.9 N vs. ~1.9–2.3 N; Figure 9). Because PE was defined as the actual minus predicted 95th-percentile force magnitude, the predominantly positive PE values indicate that the model generally underestimated peak force magnitudes. A similar but smaller pattern was found for $F_y$ PE (Figure A.10).

Comparable benefits of multiple synchronized views have been reported by others for human pose estimation, specifically that using multiple cameras yielded more accurate localization of body landmarks, particularly under occlusion or unfavorable viewing angles (Qiu et al., 2019), and for video-based kinetics, where multiple cameras improved force estimates over a single view (Uhlrich et al., 2023). In our earlier work (Rajabi et al., 2026), we estimated horizontal and vertical hand distances using the same camera infrastructure and found a considerably larger benefit of multiple views. For example, maximum absolute error for vertical distance decreased from ~32 cm with a single view to ~7 cm with three cameras. This difference may reflect underlying dependencies: distance estimation requires precise spatial localization and is therefore particularly sensitive to occlusion and monocular depth ambiguity (Cheng et al., 2019; Mehta et al., 2017; Pavllo et al., 2018), whereas the present force model relies on region-level representations of participant and handled-object motion and may therefore be less sensitive to exact landmark localization. These results suggest that multi-camera capture provides more stable estimation of peak force magnitudes, though the RMSE benefit was more modest. Thus, *Camera View Condition* appears to influence peak-force fidelity more strongly than average frame-level accuracy.

### 4.4 Effects of ROI Representation Strategy

Including the handled object as a second ROI (S2, S4) reduced RMSE relative to the participant-only strategies (S1, S3), mainly for $F_x$ during pushing and pulling (Tasks 4, 5) and for $F_z$ during symmetric and asymmetric lifting (Tasks 1, 2; ~5–7% lower), but with only negligible differences for $F_y$ (Figures 5 and 8; Figure A.11). The right-hand $F_x$ PE was likewise smaller with the object ROI (S1/S3: ~2.9 N vs. S2/S4: ~2.7 N), and for $F_z$ PE this benefit was primarily found under single-camera conditions (S2/S4: ~20–26% smaller PE vs. S1/S3), with little difference once multiple views were combined (Table A.1; Figure A.13). We suspect that including the handled box as a separate ROI provided complementary object-specific visual information that was not fully represented by the participant ROI alone. Related work has similarly emphasized the value of explicit modeling of human–object interaction for visual force and contact estimation (Pham et al., 2015, 2018; Tripathi et al., 2023). Notably, the benefit of including the object ROI here was largest under single-camera conditions, suggesting that *explicit object information may partially compensate for limited viewpoint information*.

Pixel-level segmentation, however, added little beyond detection-only ROIs (Figure 2; Figures 5 and 8; Figure A.11). Right-hand $F_x$ PE and $F_y$ RMSE, for example, were nearly identical regardless of whether the participant and object were represented as bounding boxes (S1, S2) or as segmentation masks (S3, S4). This contrasts with our earlier noted distance-estimation work (Rajabi et al., 2026), in which segmentation reduced error by ~20–30% for horizontal distance and ~35–40% for vertical distance vs. detection-only ROIs. This difference may again reflect the information required for each outcome: distance estimation depends strongly on precise spatial boundaries, whereas force estimation may depend more on the presence and relative motion of task-relevant regions than on their pixel-level extent. Thus, for force estimation, *selecting informative ROIs appears to be more important than refining those ROIs to pixel-level boundaries*.

### 4.5 Practical Implications for VLM-Based Ergonomic Assessment

These findings have several implications for the practical use of VLM-based systems in occupational physical exposure assessment. First, our results demonstrate that dynamic, triaxial, bilateral external hand forces can be estimated from RGB video and known box mass without requiring sensors on the worker or handled object as model inputs. Such estimates could provide inputs for biomechanical models for physical exposure assessment during MMH tasks while reducing the instrumentation burden associated with direct force measurement. Doing so may be particularly useful in settings where direct force measurement is impractical (e.g., objects being assembled cannot be instrumented). Second, no single *Camera View Condition* or *ROI Strategy* minimized error across all MMH tasks and force channels, suggesting that system configuration should be selected according to the task and force quantity of interest (Table 1). Including the handled object as a second ROI generally provided the most consistent benefit, particularly for pushing and pulling (Figures 5 and 8), and may therefore be a useful default when object localization is feasible. In contrast, pixel-level segmentation provided little additional improvement, suggesting that object localization may provide greater practical benefit than more computationally intensive pixel-level segmentation. Third, multi-camera configurations provided their clearest benefit for PE, particularly for $F_z$ (Figure 9). Thus, applications requiring more accurate characterization of peak force magnitudes may benefit from multiple synchronized views, whereas single-camera configurations may offer a more practical trade-off between estimation performance and system complexity when overall frame-level accuracy is sufficient. Camera configuration can therefore be selected according to the exposure metric of primary interest and the practical constraints of the work environment.

### 4.6 Limitations and Future Research Directions

Several limitations of our study should be acknowledged. All participants were relatively young (18–39 years old) and healthy, and all tasks were performed under controlled laboratory conditions using a single box and relatively consistent lighting and background. In addition, although LOSO evaluated generalization to unseen participants, the same laboratory setup and three camera viewpoints were used across folds. Thus, generalization to unseen camera viewpoints, substantially different camera placements, or a new visual environment was not evaluated. It thus remains unknown regarding the extent to which the present results generalize to occupationally experienced or more diverse worker populations, different handled objects, and workplace conditions involving clutter, variable lighting, and occlusion. Future work should evaluate performance under progressively less controlled and ultimately real occupational conditions. Second, our DINOv2 features represented task-relevant regions at the ROI level without explicitly representing hand, joint, or hand–object contact geometry. The task-dependent benefit of including the handled object as a second ROI suggests that more explicit representations of participant–object interaction may provide additional information for force estimation. Prior work has reported benefits from incorporating hand pose or body–object contact information when estimating interaction quantities from visual data (Tripathi et al., 2023; Zhao et al., 2025). Future models could therefore augment ROI features with hand or body keypoints, explicit object motion, or learned contact representations.

Third, the transformer operated on fixed, 300-frame sequences using bidirectional temporal context, and all processing was performed offline. While PE characterized differences in 95th-percentile force magnitude, it did not assess whether high-force portions occurred at the same time in the actual and predicted force signals. This temporal aspect may be important when characterizing dynamic physical exposures. Although real-time estimation is not necessary for many occupational exposure or risk assessment applications, the computational demands of applying the pipeline to longer or continuously recorded workplace videos were not evaluated. Future work should therefore assess the temporal alignment of high-force events, processing efficiency, and the effects of different temporal window lengths when applying our approach to larger workplace video datasets. Finally, the present dataset involved one box geometry and three known mass levels (6, 9, and 12 kg), with box mass explicitly provided to the prediction model. Generalization to objects with different shapes, dimensions, mass distributions, or unknown masses was therefore not evaluated. Testing across a wider range of handled objects, and determining how load information can be provided or estimated in practical applications, will be important for broader deployment.

## 5.0 Conclusions

Accurately and non-invasively estimating dynamic, triaxial, bilateral external hand forces during MMH tasks is an important step toward practical, video-based occupational physical exposure assessment. We developed and evaluated a multi-stage VLM-based pipeline integrating text-guided ROI detection and segmentation, pretrained vision-transformer feature extraction, and transformer-based temporal regression across five *MMH tasks*, seven *Camera View Conditions*, and four *ROI Strategies*. Overall, RMSE was ~4.7–5.6 N for the horizontal and mediolateral hand force components and ~10.6–11.0 N for the vertical component, demonstrating the feasibility of estimating detailed external hand forces using RGB video and known box mass as model inputs. Estimation performance varied with *MMH Task*, *Mass*, *Camera View Condition*, and *ROI Strategy*. Including the handled object as a second ROI generally improved force estimation and, for some outcomes, provided the greatest benefit under single-camera conditions, suggesting that explicit object information may partially compensate for limited viewpoint information. Pixel-level segmentation, in contrast, provided little additional benefit, suggesting that selecting informative ROIs may be more important than refining them to pixel-level boundaries. Multi-camera capture produced the clearest advantage for peak-force estimation, while differences in overall RMSE among camera conditions were comparatively modest. Together, these findings demonstrate that detailed, continuous external hand-force estimates can be obtained from RGB video with substantially reduced instrumentation compared with conventional direct or model-based approaches. However, further research is needed to evaluate performance across diverse worker populations and handled objects, and in real occupational settings.

## 6.0 Declaration of generative AI and AI-assisted technologies in the manuscript preparation process

During the preparation of this work, we used ChatGPT to refine some sentences and improve the clarity of the text. After using this tool/service, we reviewed and edited the content as needed, and we take full responsibility for the content of this publication.

## 7.0 Acknowledgements

The first author was supported by a predoctoral training program grant (T03 OH008613) from CDC/NIOSH. The current contents are solely the authors' responsibility and do not necessarily represent the official views of NIOSH or the CDC. The authors thank Advanced Research Computing at Virginia Tech for providing computational resources and technical support that contributed to the results reported in this paper. URL: https://arc.vt.edu/

## References

Bezzini, R., Crosato, L., Teppati Losè, M., Avizzano, C. A., Bergamasco, M., & Filippeschi, A. (2023). Closed-Chain Inverse Dynamics for the Biomechanical Analysis of Manual Material Handling Tasks through a Deep Learning Assisted Wearable Sensor Network. Sensors, 23(13), 5885. https://doi.org/10.3390/s23135885

Blasco-Abadía, J., Bellosta-López, P., Doménech-García, V., Palsson, T. S., Christensen, S. W. M., Hoegh, M., Berjano, P., & Langella, F. (2025). Cross-cultural adaptation and validation of the Spanish version of the Prevent for Work questionnaire. Frontiers in Public Health, 12, 1453492. https://doi.org/10.3389/fpubh.2024.1453492

Burdorf, A. (1995). Reducing random measurement error in assessing postural load on the back in epidemiologic surveys. Scandinavian Journal of Work, Environment & Health, 21(1), 15–23.

Burdorf, A., & van der Beek, A. (1999). Exposure assessment strategies for work-related risk factors for musculoskeletal disorders. Scandinavian Journal of Work, Environment & Health, 25, 25–30.

Carreira, J., & Zisserman, A. (2017). Quo Vadis, Action Recognition? A New Model and the Kinetics Dataset (Version 3). arXiv. https://doi.org/10.48550/ARXIV.1705.07750

Cheng, Y., Yang, B., Wang, B., Yan, W., & Tan, R. T. (2019). Occlusion-aware networks for 3d human pose estimation in video. Proceedings of the IEEE/CVF International Conference on Computer Vision, 723–732.

Da Costa, B. R., & Vieira, E. R. (2010). Risk factors for work-related musculoskeletal disorders: A systematic review of recent longitudinal studies. American Journal of Industrial Medicine, 53(3), 285–323. https://doi.org/10.1002/ajim.20750

Dempsey, P. G., McGorry, R. W., & Maynard, W. S. (2005). A survey of tools and methods used by certified professional ergonomists. Applied Ergonomics, 36(4), 489–503. https://doi.org/10.1016/j.apergo.2005.01.007

Deng, J., Dong, W., Socher, R., Li, L.-J., Kai Li, & Li Fei-Fei. (2009). ImageNet: A large-scale hierarchical image database. 2009 IEEE Conference on Computer Vision and Pattern Recognition, 248–255. https://doi.org/10.1109/CVPR.2009.5206848

Faber, G. S., Chang, C.-C., Kingma, I., & Dennerlein, J. T. (2013). Estimating dynamic external hand forces during manual materials handling based on ground reaction forces and body segment accelerations. Journal of Biomechanics, 46(15), 2736–2740. https://doi.org/10.1016/j.jbiomech.2013.07.030

Faber, G. S., Koopman, A. S., Kingma, I., Chang, C. C., Dennerlein, J. T., & Van Dieën, J. H. (2018). Continuous ambulatory hand force monitoring during manual materials handling using instrumented force shoes and an inertial motion capture suit. Journal of Biomechanics, 70, 235–241. https://doi.org/10.1016/j.jbiomech.2017.10.006

Garosi, E., Sheikh, F., & Goodarzi, M. (2025). Ergonomic Interventions in Risk Reduction. In H. Gül (Ed.), Public Health (Vol. 3). IntechOpen. https://doi.org/10.5772/intechopen.1008463

Gholamiangonabadi, D., Kiselov, N., & Grolinger, K. (2020). Deep Neural Networks for Human Activity Recognition With Wearable Sensors: Leave-One-Subject-Out Cross-Validation for Model Selection. IEEE Access, 8, 133982–133994. https://doi.org/10.1109/ACCESS.2020.3010715

Grady, P., Tang, C., Brahmbhatt, S., Twigg, C. D., Wan, C., Hays, J., & Kemp, C. C. (2022). PressureVision: Estimating hand pressure from a single RGB image. European Conference on Computer Vision, 328–345.

Hendrycks, D., & Gimpel, K. (2016). Gaussian Error Linear Units (GELUs) (Version 5). arXiv. https://doi.org/10.48550/ARXIV.1606.08415

Hlucny, S. D., & Novak, D. (2020). Characterizing Human Box-Lifting Behavior Using Wearable Inertial Motion Sensors. Sensors, 20(8), 2323. https://doi.org/10.3390/s20082323

Javanmardi, S., Rappelt, L., Baumgart, C., Niederer, D., Heinke, L., & Freiwald, J. (2025). Work conditions and determinants of health status among industrial shift workers: A cross-sectional study. Frontiers in Public Health, 12, 1489178. https://doi.org/10.3389/fpubh.2024.1489178

Jordao, A., Nazare, A. C., Sena, J., & Schwartz, W. R. (2018). Human Activity Recognition Based on Wearable Sensor Data: A Standardization of the State-of-the-Art (Version 3). arXiv. https://doi.org/10.48550/ARXIV.1806.05226

Kang, D., Jeong, D., Lee, H., Park, S., Park, H., Kwon, S., Kim, Y., & Paik, J. (2025). VLM-HOI: Vision Language Models for Interpretable Human-Object Interaction Analysis. In A. Del Bue, C. Canton, J. Pont-Tuset, & T. Tommasi (Eds.), Computer Vision – ECCV 2024 Workshops

(Vol. 15634, pp. 218–235). Springer Nature Switzerland. https://doi.org/10.1007/978-3-031-92591-7_14

Kirillov, A., Mintun, E., Ravi, N., Mao, H., Rolland, C., Gustafson, L., Xiao, T., Whitehead, S., Berg, A. C., Lo, W.-Y., Dollár, P., & Girshick, R. (2023). Segment Anything (arXiv:2304.02643). arXiv. https://doi.org/10.48550/arXiv.2304.02643

Koppelaar, E., & Wells, R. (2005). Comparison of measurement methods for quantifying hand force. Ergonomics, 48(8), 983–1007. https://doi.org/10.1080/00140130500120841

Kubota, K., Aoki, H., Nakamura, A., Suzuki, R., & Kanemura, N. (2026). Force-plate-free estimation of ground reaction forces and lower-limb joint moments during gait using a single RGB-D camera and physics-based optimization. Gait & Posture, 130, 110285. https://doi.org/10.1016/j.gaitpost.2026.110285

Kumar, S. (1993). Perception of posture of short duration in the spatial and temporal domains. Applied Ergonomics, 24(5), 345–350. https://doi.org/10.1016/0003-6870(93)90073-I

Lan, Y., Zhan, X., Wang, Z., Jiang, D., Li, X., & Peng, C. (2026). Global prevalence and associated risk factors of work-related musculoskeletal disorders among steelworkers: A systematic review and meta-analysis. Frontiers in Public Health, 14, 1718101. https://doi.org/10.3389/fpubh.2026.1718101

Lee, H., Yang, K., Kim, N., & Ahn, C. R. (2020). Detecting excessive load-carrying tasks using a deep learning network with a Gramian Angular Field. Automation in Construction, 120, 103390. https://doi.org/10.1016/j.autcon.2020.103390

Lee, T.-H. (2015). The effects of load magnitude and lifting speed on the kinematic data of load and human posture. International Journal of Occupational Safety and Ergonomics, 21(1), 55–61. https://doi.org/10.1080/10803548.2015.1017956

Liberty Mutual Insurance. (2023). 2023 Workplace Safety Index. 2023 Workplace Safety Index: The Top 10 Causes of Disabling Injuries. https://business.libertymutual.com/insights/2023-workplace-safety-index/

Lim, S. (2024). Exposures to select risk factors can be estimated from a continuous stream of inertial sensor measurements during a variety of lifting-lowering tasks. Ergonomics, 1–16. https://doi.org/10.1080/00140139.2024.2343949

Lim, S., & D’Souza, C. (2019). Statistical prediction of load carriage mode and magnitude from inertial sensor derived gait kinematics. Applied Ergonomics, 76, 1–11. https://doi.org/10.1016/j.apergo.2018.11.007

Liu, S., Zeng, Z., Ren, T., Li, F., Zhang, H., Yang, J., Jiang, Q., Li, C., Yang, J., Su, H., Zhu, J., & Zhang, L. (2024). Grounding DINO: Marrying DINO with Grounded Pre-Training for Open-Set Object Detection (arXiv:2303.05499). arXiv. https://doi.org/10.48550/arXiv.2303.05499

Louis, N., Corso, J. J., Templin, T. N., Eliason, T. D., & Nicolella, D. P. (2022). Learning to Estimate External Forces of Human Motion in Video. Proceedings of the 30th ACM International Conference on Multimedia, MM ’22, 3540–3548. https://doi.org/10.1145/3503161.3548377

Mathiassen, S. E., & Winkel, J. (1991). Quantifying variation in physical load using exposure-vs-time data. Ergonomics, 34(12), 1455–1468. https://doi.org/10.1080/00140139108964889

Mehta, D., Sridhar, S., Sotnychenko, O., Rhodin, H., Shafiei, M., Seidel, H.-P., Xu, W., Casas, D., & Theobalt, C. (2017). VNect: Real-time 3D human pose estimation with a single RGB camera. ACM Transactions on Graphics, 36(4), 1–14. https://doi.org/10.1145/3072959.3073596

Mobasser, F., & Hashtrudi-Zaad, K. (2012). A Comparative Approach to Hand Force Estimation using Artificial Neural Networks. Biomedical Engineering and Computational Biology, 4, BECB.S9335. https://doi.org/10.4137/BECB.S9335

Ojelade, A. E. (2024). Evaluation of Markerless Motion Capture to Assess Physical Exposures During Material Handling Tasks.

Ojelade, A., Rajabi, M. S., Kim, S., & Nussbaum, M. A. (2025). A Data-Driven Approach to Classifying Manual Material Handling Tasks Using Markerless Motion Capture and Recurrent Neural Networks. International Journal of Industrial Ergonomics. https://doi.org/10.1016/j.ergon.2025.103755

Oquab, M., Darcet, T., Moutakanni, T., Vo, H., Szafraniec, M., Khalidov, V., Fernandez, P., Haziza, D., Massa, F., El-Nouby, A., Assran, M., Ballas, N., Galuba, W., Howes, R., Huang, P.-Y., Li, S.-W., Misra, I., Rabbat, M., Sharma, V., … Bojanowski, P. (2024). DINOv2: Learning Robust Visual Features without Supervision (arXiv:2304.07193). arXiv. https://doi.org/10.48550/arXiv.2304.07193

Paudel, P., Kwon, Y.-J., Kim, D.-H., & Choi, K.-H. (2022). Industrial Ergonomics Risk Analysis Based on 3D-Human Pose Estimation. Electronics, 11(20), 3403. https://doi.org/10.3390/electronics11203403

Pavllo, D., Feichtenhofer, C., Grangier, D., & Auli, M. (2018). 3D human pose estimation in video with temporal convolutions and semi-supervised training (Version 2). arXiv. https://doi.org/10.48550/ARXIV.1811.11742

Pham, T.-H., Kheddar, A., Qammaz, A., & Argyros, A. A. (2015). Towards force sensing from vision: Observing hand-object interactions to infer manipulation forces. 2015 IEEE Conference on Computer Vision and Pattern Recognition (CVPR), 2810–2819. https://doi.org/10.1109/CVPR.2015.7298898

Pham, T.-H., Kyriazis, N., Argyros, A. A., & Kheddar, A. (2018). Hand-Object Contact Force Estimation from Markerless Visual Tracking. IEEE Transactions on Pattern Analysis and Machine Intelligence, 40(12), 2883–2896. https://doi.org/10.1109/TPAMI.2017.2759736

Porta, M., Kim, S., Pau, M., & Nussbaum, M. A. (2021). Classifying diverse manual material handling tasks using a single wearable sensor. Applied Ergonomics, 93, 103386. https://doi.org/10.1016/j.apergo.2021.103386

Prechelt, L. (1998). Early Stopping—But When? In G. B. Orr & K.-R. Müller (Eds.), Neural Networks: Tricks of the Trade (pp. 55–69). Springer. https://doi.org/10.1007/3-540-49430-8_3

Qiu, H., Wang, C., Wang, J., Wang, N., & Zeng, W. (2019). Cross view fusion for 3d human pose estimation. Proceedings of the IEEE/CVF International Conference on Computer Vision, 4342–4351.

Radford, A., Kim, J. W., Hallacy, C., Ramesh, A., Goh, G., Agarwal, S., Sastry, G., Askell, A., Mishkin, P., Clark, J., Krueger, G., & Sutskever, I. (2021). Learning Transferable Visual Models From Natural Language Supervision (arXiv:2103.00020). arXiv. https://doi.org/10.48550/arXiv.2103.00020

Rajabi, M. S., Ojelade, A., Kim, S., & Nussbaum, M. A. (2026). Vision-Language Models for Ergonomic Assessment of Manual Lifting Tasks: Estimating Horizontal and Vertical Hand Distances from RGB Video (arXiv:2602.20658). arXiv. https://doi.org/10.48550/arXiv.2602.20658

Rajabi, M. S., Ojelade, A., Kim, S., & Nussbaum, M. A. (2027). Vision-language models for occupational physical exposure assessment: Classification and temporal segmentation of manual material handling tasks. Applied Ergonomics, 138, 104831. https://doi.org/10.1016/j.apergo.2026.104831

Saponas, T. S., Tan, D. S., Morris, D., & Balakrishnan, R. (2008). Demonstrating the feasibility of using forearm electromyography for muscle-computer interfaces. Proceedings of the SIGCHI Conference on Human Factors in Computing Systems, 515–524. https://doi.org/10.1145/1357054.1357138

Sigal, L., Balan, A. O., & Black, M. J. (2010). HumanEva: Synchronized Video and Motion Capture Dataset and Baseline Algorithm for Evaluation of Articulated Human Motion. International Journal of Computer Vision, 87(1–2), 4–27. https://doi.org/10.1007/s11263-009-0273-6

Smith, M., Madinei, S., Livingston, J., & Yu, A. (2024). User Expectations for Comfort, Discomfort, and Wearability of Wrist-Worn Devices. Proceedings of the Human Factors and Ergonomics Society Annual Meeting, 68(1), 689–692. https://doi.org/10.1177/10711813241260673

Spielholz, P., Silverstein, B., Morgan, M., Checkoway, H., & Kaufman, J. (2001). Comparison of self-report, video observation and direct measurement methods for upper extremity musculoskeletal disorder physical risk factors. Ergonomics, 44(6), 588–613. https://doi.org/10.1080/00140130118050

Steinebach, T., Grosse, E. H., Glock, C. H., Wakula, J., & Lunin, A. (2020). Accuracy evaluation of two markerless motion capture systems for measurement of upper extremities: Kinect V2 and Captiv. Human Factors and Ergonomics in Manufacturing & Service Industries, 30(4), 291–302. https://doi.org/10.1002/hfm.20840

Tang, W., Shao, L., & Chen, X. (2025). Estimation of Hand Pressure and Pose From RGB Images Based on Cross-Modal Cues. IEEE Sensors Journal, 25(1), 2030–2039. https://doi.org/10.1109/JSEN.2024.3502449

Taori, S., & Lim, S. (2024). Use of a wearable electromyography armband to detect lift-lower tasks and classify hand loads. Applied Ergonomics, 119, 104285. https://doi.org/10.1016/j.apergo.2024.104285

Tripathi, S., Chatterjee, A., Passy, J.-C., Yi, H., Tzionas, D., & Black, M. J. (2023). DECO: Dense Estimation of 3D Human-Scene Contact In The Wild. 2023 IEEE/CVF International

Conference on Computer Vision (ICCV), 7967–7979. https://doi.org/10.1109/ICCV51070.2023.00735

Uhlrich, S. D., Falisse, A., Kidziński, Ł., Muccini, J., Ko, M., Chaudhari, A. S., Hicks, J. L., & Delp, S. L. (2023). OpenCap: Human movement dynamics from smartphone videos. PLOS Computational Biology, 19(10), e1011462. https://doi.org/10.1371/journal.pcbi.1011462

U.S. Bureau of Labor Statistics. (2024). Survey of Occupational Injuries and Illnesses Data. Nonfatal Occupational Injuries and Illnesses Requiring Days Away from Work. https://www.bls.gov/data/home.htm

Van Der Beek, A. J., & Frings-Dresen, M. H. (1998). Assessment of mechanical exposure in ergonomic epidemiology. Occupational and Environmental Medicine, 55(5), 291–299. https://doi.org/10.1136/oem.55.5.291

Vaswani, A., Shazeer, N., Parmar, N., Uszkoreit, J., Jones, L., Gomez, A. N., Kaiser, Ł., & Polosukhin, I. (2017). Attention is all you need. Advances in Neural Information Processing Systems, 30.

Wade, L., Needham, L., McGuigan, P., & Bilzon, J. (2022). Applications and limitations of current markerless motion capture methods for clinical gait biomechanics. PeerJ, 10, e12995. https://doi.org/10.7717/peerj.12995

Wang, M., Zhao, C., Barr, A., Fan, H., Yu, S., Kapellusch, J., & Harris Adamson, C. (2023). Hand Posture and Force Estimation Using Surface Electromyography and an Artificial Neural Network. Human Factors: The Journal of the Human Factors and Ergonomics Society, 65(3), 382–402. https://doi.org/10.1177/00187208211016695

Wells, R., Norman, R., Neumann, P., Andrews, D., Frank, J., Shannon, H., & Kerr, M. (1997). Assessment of physical work load in epidemiologic studies: Common measurement metrics for exposure assessment. Ergonomics, 40(1), 51–61. https://doi.org/10.1080/001401397188369

Wells, R., Van Eerd, D., & Hägg, G. (2004). Mechanical exposure concepts using force as the agent. Scandinavian Journal of Work, Environment & Health, 30(3), 179–190.

Wiktorin, C., Hjelm, E. W., Winkel, J., Köster, M., & Group, %Stockholm MUSIC I. Study. (1996). Reproducibility of a Questionnaire for Assessment of Physical Load During Work and Leisure Time. Journal of Occupational and Environmental Medicine, 38(2), 190.

Wiktorin, C., Karlqvist, L., Winkel, J., & group, S. M. I. study. (1993). Validity of self-reported exposures to work postures and manual materials handling. Scandinavian Journal of Work, Environment & Health, 19(3), 208–214.

Winkel, J., & Mathiassen, S. E. (1994). Assessment of physical work load in epidemiologic studies: Concepts, issues and operational considerations. Ergonomics, 37(6), 979–988. https://doi.org/10.1080/00140139408963711

Yang, H., Haldeman, S., Lu, M.-L., & Baker, D. (2016). Low Back Pain Prevalence and Related Workplace Psychosocial Risk Factors: A Study Using Data From the 2010 National Health Interview Survey. Journal of Manipulative and Physiological Therapeutics, 39(7), 459–472. https://doi.org/10.1016/j.jmpt.2016.07.004

Zhao, W., Yang, G., Zhang, R., Jiang, C., Yang, C., Yan, Y., Hussain, A., & Huang, K. (2023). Open-Pose 3D Zero-Shot Learning: Benchmark and Challenges (Version 2). arXiv. https://doi.org/10.48550/ARXIV.2312.07039

Zhao, Y., Kwon, T., Streli, P., Pollefeys, M., & Holz, C. (2025). Egopressure: A dataset for hand pressure and pose estimation in egocentric vision. Proceedings of the Computer Vision and Pattern Recognition Conference, 27727–27738.

Zheng, C., Zhu, S., Mendieta, M., Yang, T., Chen, C., & Ding, Z. (2021). 3D Human Pose Estimation with Spatial and Temporal Transformers (Version 3). arXiv. https://doi.org/10.48550/ARXIV.2103.10455

Zhu, Z., Mu, F., Radwin, R., & Li, Y. (2025). Towards video-based injury risk assessment: Predicting lifting loads from body pose trajectories. Machine Vision and Applications, 36(6), 136. https://doi.org/10.1007/s00138-025-01758-w

Zhuang, J., Liu, Y., Jia, Y., & Huang, Y. (2019). User Discomfort Evaluation Research on the Weight and Wearing Mode of Head-Wearable Device. In T. Z. Ahram (Ed.), Advances in Human Factors in Wearable Technologies and Game Design (Vol. 795, pp. 98–110). Springer International Publishing. https://doi.org/10.1007/978-3-319-94619-1_10

## Appendix

### A.1 Participants

Data used here were obtained from a prior study (Ojelade et al., 2025; Ojelade, 2024). A convenience sample of 35 young adults (21 males and 14 females) completed the study and was recruited from the university and local community. Respective means (SD) of age, body mass, and height were 27.2 (4.5) years, 77.6 (12.3) kg, and 176.4 (5.7) cm for the males; and 26.8 (5.5) years, 68.2 (8.2) kg, and 170.1 (7.1) cm for the females. All participants self-reported being right-handed and physically active (i.e., exercising at least twice per week), and having no musculoskeletal disorders within the past year. The research reported herein compiled with the tenets of the Declaration of Helsinki, and the study protocol was approved by the Institutional Review Board at Virginia Tech (#23-095). Informed consent was obtained from all participants prior to any data collection.

### A.2 MMH Task Details and Experimental Procedures

A single wood box (width = 26.0 cm; depth = 41.0 cm; and height = 23.5 cm; Figure A.1) was used for the five simulated MMH tasks described in the main text. A repeated-measures experimental design was implemented. Participants completed a training phase to practice the tasks using their comfortable work strategies and speed, simulating an industrial setting. During the experimental phase, the order of *Hand Configurations* and *Box Mass* presentation was counterbalanced using balanced Latin square designs to reduce potential bias. To minimize physical fatigue, a mandatory rest period of at least four minutes was provided between trials involving different hand configurations.

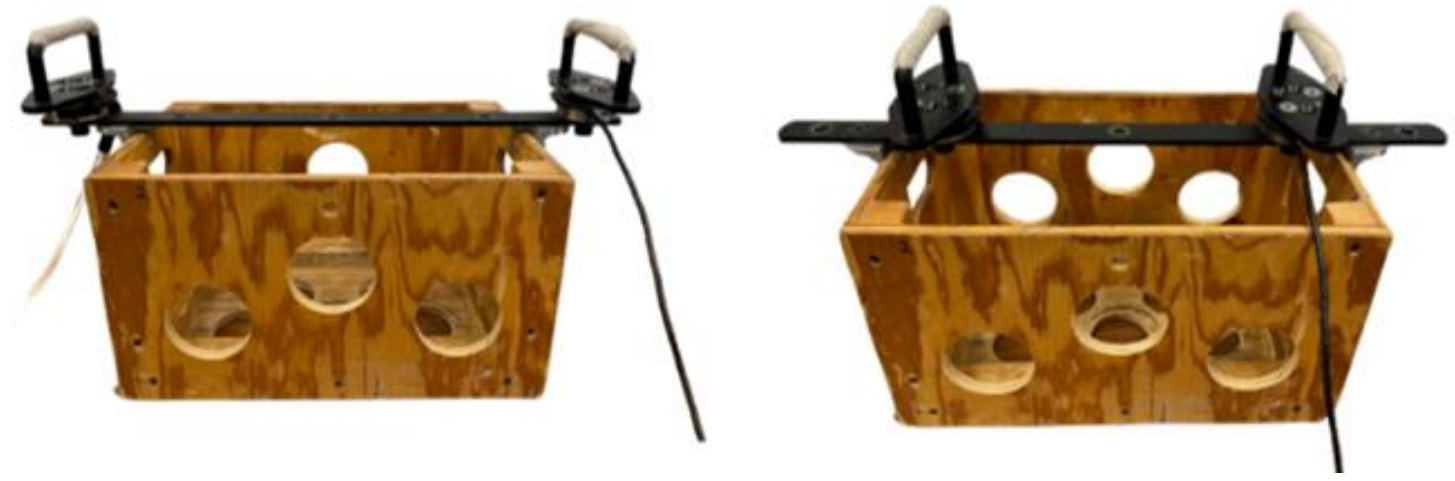

Figure A.1. Illustration of the two hand configurations: broad (left), and narrow (right; adapted from Ojelade et al. (2025)

### A.3 Azure Kinect™ Camera Instrumentation

Whole-body kinematics were recorded using three synchronized Azure Kinect™ markerless camera systems (Microsoft Corporation, Seattle, WA, USA), which were each sampled at 30 Hz. The cameras were positioned approximately 1.74 m from the edge of the work area; this configuration was established during pilot testing to improve coverage given the narrow field of view of the camera units. The three cameras were time-synchronized using a 3.5-mm auxiliary cable connected in a daisy-chain configuration, with one camera designated as the primary device and the other two as secondary devices.

### A.4 Load Cell Instrumentation and Hand Force Measurements

External hand forces were obtained using triaxial load cells (Michigan Scientific Corp. TR3D-A-1K, Charlevoix, MI, USA) attached near the top of the box. Load-cell signals were sampled at 200 Hz using a custom LabView program, filtered using a 300-ms moving root-mean-square window, and were then downsampled to 30 Hz to match the sampling rate of the kinematic data. Tri-axial hand forces were represented in a local, box-centered coordinate system (Figure A.2) and served as the ground-truth target variables for subsequent model development and evaluation.

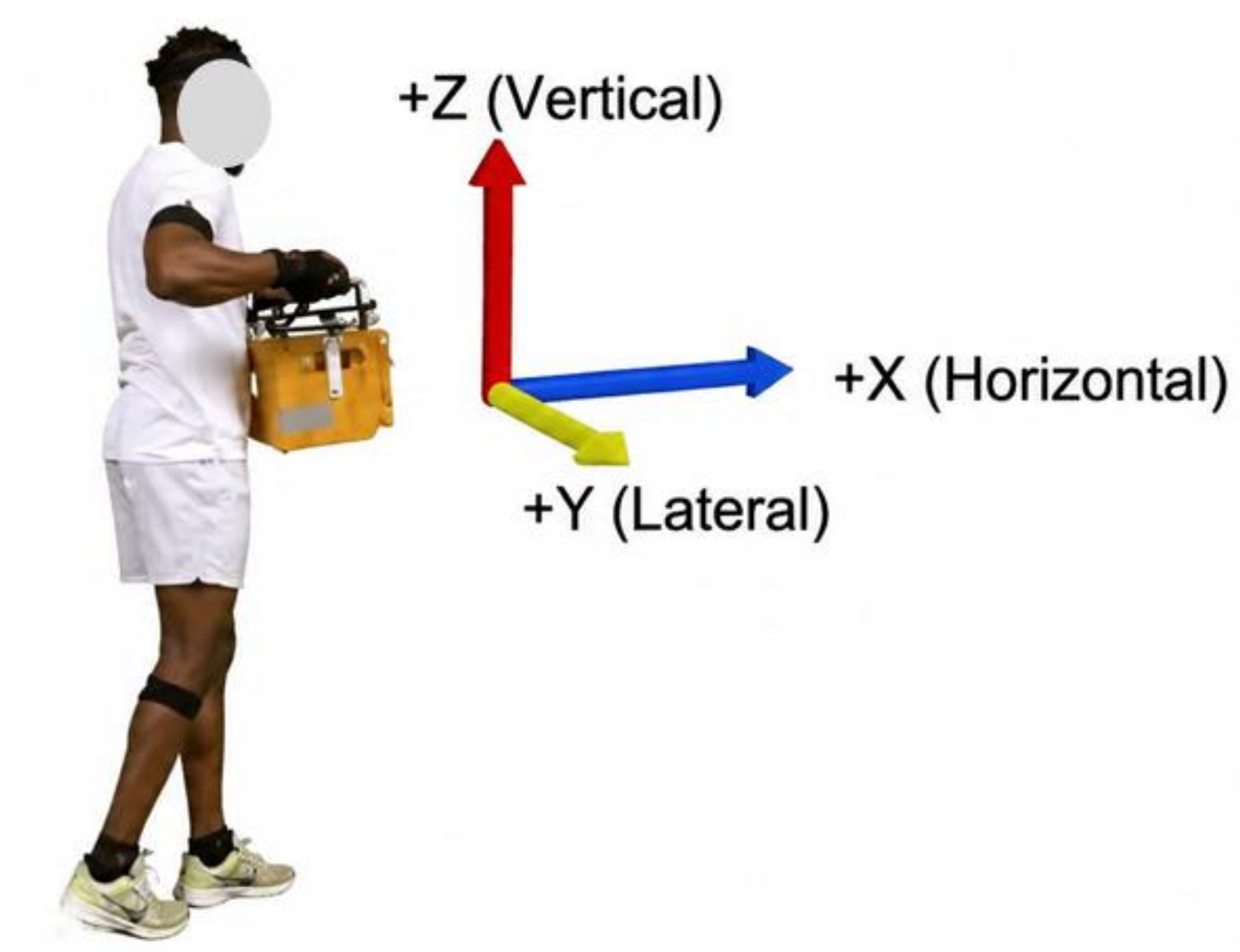

Figure A.2. Illustration of the box-centered coordinate system used to define the external hand forces during MMH tasks. The +X axis represents the horizontal/anterior–posterior direction, the +Y axis represents the lateral/medial–lateral direction, and the +Z axis represents the vertical direction.

### A.5 Transformer-Based Force Prediction Model: Additional Details

We used a modified transformer encoder model with the following configuration: latent dimension d_model = 256; number of attention heads = 8; number of transformer encoder layers = 4; feedforward sublayer dimension = 512; dropout = 0.1. The regression head consisted of a linear layer (256 → 128 units) with GELU activation and dropout, followed by a linear layer (128 → 6 units) producing simultaneous per-frame estimates of all six force channels. Model weights were initialized using Xavier uniform initialization for linear layers, normal initialization (standard deviation = 0.02) for embedding layers, and ones and zeros for layer normalization weight and bias parameters, respectively.

We used a fixed-length windows of 300 frames were extracted from each trial sequence using a sliding-window approach. For training, windows were generated with a 150-frame stride (50% overlap) to increase the number of training samples and improve temporal generalization. For validation and test sets, non-overlapping windows (stride = 300) were used to avoid redundant evaluation across frames. If a trial was shorter than 300 frames, a single zero-padded window covering the entire trial was created, with an attention mask set to False for all padded positions.

This method ensured that padded frames did not contribute to either the loss computation or the self-attention mechanism during encoding.

The AdamW optimizer was used with an initial learning rate of $1 \times 10^{-4}$, weight decay of $1 \times 10^{-4}$, and gradient clipping at a maximum norm of 1.0. The learning rate was annealed using a cosine annealing schedule over 100 epochs, with a minimum learning rate of 1% of the initial value. The batch size was set to 16. Training was performed with early stopping applied when validation loss failed to improve over 20 consecutive epochs; the model checkpoint corresponding to the minimum validation loss within a fold was loaded for final evaluation on that fold's test participant. We used the following evaluation metrics:

$$\textit{Mean Absolute Error (MAE)} = \frac{1}{N}\sum_{i=1}^{N} |\, \hat{y_i} - y_i \,|$$

$$\textit{Root Mean Square Error (RMSE)} = \sqrt{\frac{1}{N}\sum_{i=1}^{N} (\hat{y}_i - y_i)^2}$$

$$\textit{Peak Error (PE)} = \mathrm{p}_{95}(|\, \mathrm{y} \,|) - \mathrm{p}_{95}(|\, \hat{\mathrm{y}} \,|)$$

where $\hat{y}_i$ and $y_i$ are the predicted and ground-truth force values, respectively, $N$ is the number of valid (non-padded) frames in the test set for a given fold, and $\mathrm{p}_{95}(\cdot)$ denotes the 95th percentile of the absolute values across all valid frames. Each metric was computed separately for each of the six bilateral force channels.

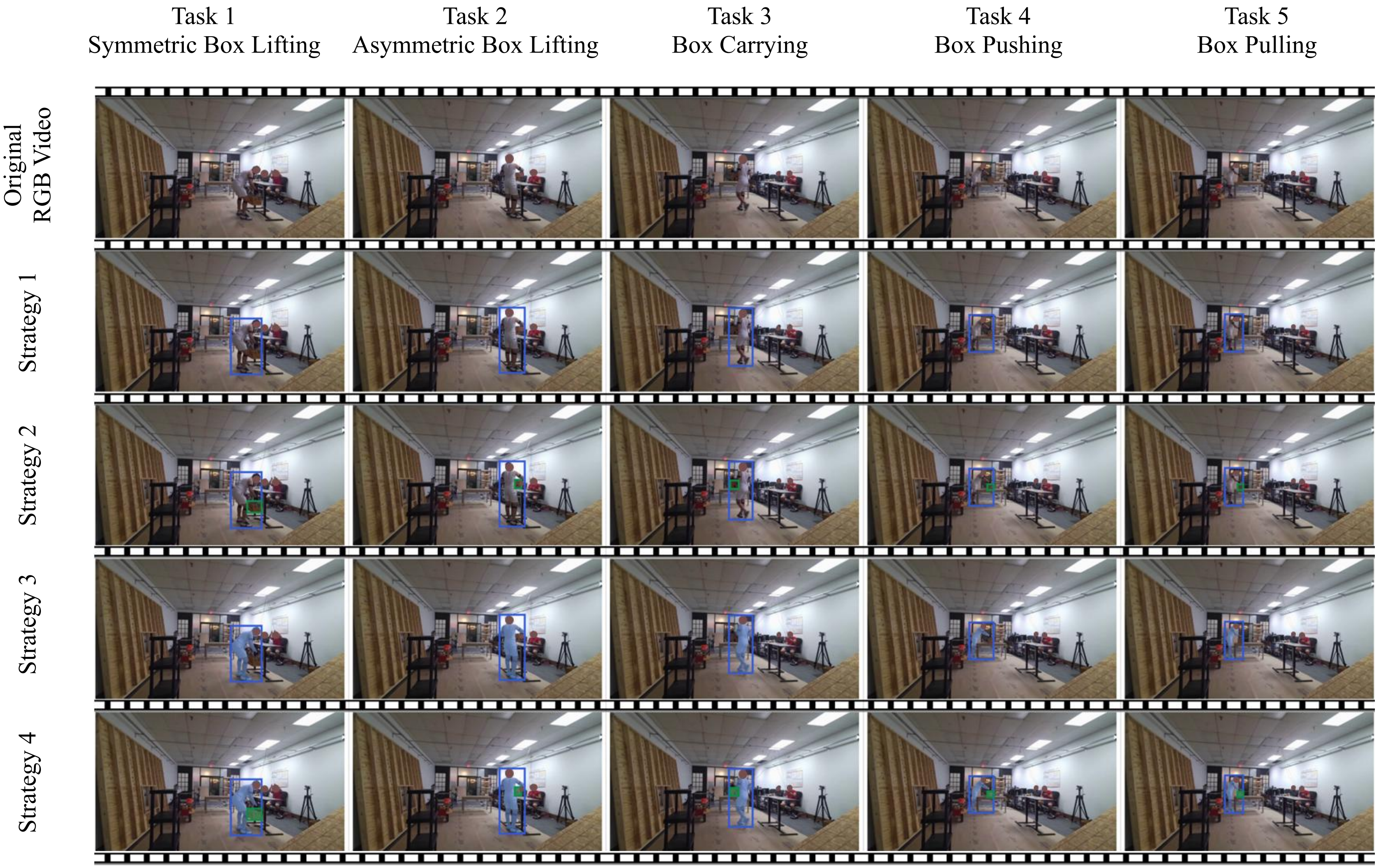

Figure A.3. Example frames from View 2 (V2) illustrating the five MMH tasks and the four ROI representation strategies evaluated in this study. The first row shows the original RGB frames extracted from the video recordings. Subsequent rows illustrate the four ROI processing strategies: Strategy 1, participant bounding box; Strategy 2, participant and handled-object bounding boxes; Strategy 3, participant segmentation mask; and Strategy 4, participant and handled-object segmentation masks. Participant regions are shown in blue and handled-object regions in green.

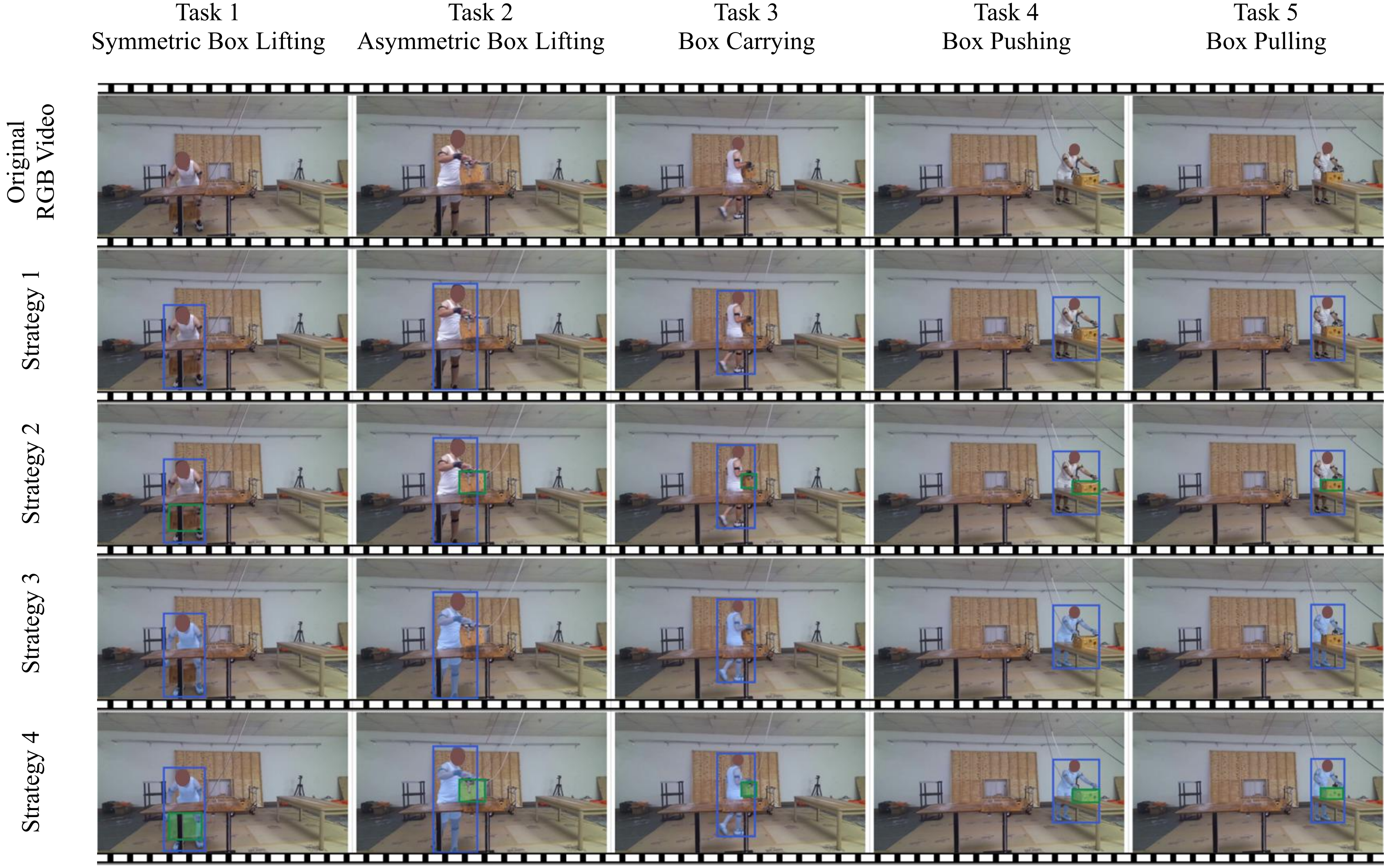

Figure A.4. Example frames from View 3 (V3) illustrating the five MMH tasks and the four ROI representation strategies evaluated in this study. The first row shows the original RGB frames extracted from the video recordings. Subsequent rows illustrate the four ROI processing strategies: Strategy 1, participant bounding box; Strategy 2, participant and handled-object bounding boxes; Strategy 3, participant segmentation mask; and Strategy 4, participant and handled-object segmentation masks. Participant regions are shown in blue and handled-object regions in green.

Table A.1. LMM results assessing the effects of *Biological Sex*, *Hand Configuration*, *Mass*, *Camera View Condition*, *MMH Task*, and *ROI Strategy* on bilateral **horizontal force ($F_x$)** prediction performance metrics. Entries are *F* values (*p* values), and significant effects are highlighted in **bold** font.

| Effect | MAE (Right Hand) | MAE (Left Hand) | RMSE (Right Hand) | RMSE (Left Hand) | Peak Error (Right Hand) | Peak Error (Left Hand) |
|---|---|---|---|---|---|---|
| *Biological Sex (B)* | 0.33 (0.5708) | **7.61 (0.0094)** | 0.34 (0.5613) | **7.45 (0.0101)** | 0.44 (0.5101) | **4.97 (0.0328)** |
| *Hand Configuration (H)* | **31.61 (<.0001)** | **5.60 (0.0180)** | **30.06 (<.0001)** | **3.90 (0.0483)** | **10.94 (0.0009)** | 3.68 (0.0550) |
| *Mass (M)* | **2746.73 (<.0001)** | **3349.84 (<.0001)** | **3089.90 (<.0001)** | **3719.66 (<.0001)** | **939.56 (<.0001)** | **1104.60 (<.0001)** |
| *Camera View Condition (C)* | 0.83 (0.5495) | 1.16 (0.3242) | **2.51 (0.0200)** | **2.25 (0.0354)** | **21.17 (<.0001)** | **20.20 (<.0001)** |
| *MMH Task (T)* | **10949.58 (<.0001)** | **15881.23 (<.0001)** | **12747.18 (<.0001)** | **17193.10 (<.0001)** | **420.67 (<.0001)** | **1627.70 (<.0001)** |
| *ROI Strategy (S)* | **6.49 (0.0002)** | **9.27 (<.0001)** | **6.30 (0.0003)** | **8.04 (<.0001)** | **6.83 (0.0001)** | **33.93 (<.0001)** |
| B × H | **20.62 (<.0001)** | 2.99 (0.0840) | **20.60 (<.0001)** | 2.83 (0.0925) | 0.69 (0.4075) | 1.52 (0.2179) |
| B × M | **76.56 (<.0001)** | **36.26 (<.0001)** | **85.53 (<.0001)** | **40.71 (<.0001)** | **52.37 (<.0001)** | **17.49 (<.0001)** |
| B × C | 0.61 (0.7196) | **2.31 (0.0310)** | 0.83 (0.5495) | **2.43 (0.0241)** | **3.07 (0.0052)** | 1.99 (0.0635) |
| B × T | **39.21 (<.0001)** | **640.57 (<.0001)** | **42.13 (<.0001)** | **645.86 (<.0001)** | **31.56 (<.0001)** | **253.19 (<.0001)** |
| B × S | **8.24 (<.0001)** | **3.11 (0.0251)** | **8.41 (<.0001)** | **3.27 (0.0204)** | 2.09 (0.0988) | **4.89 (0.0021)** |
| H × M | **8.04 (0.0003)** | **7.56 (0.0005)** | **8.58 (0.0002)** | **11.35 (<.0001)** | **10.34 (<.0001)** | **4.72 (0.0089)** |
| H × C | 0.43 (0.8561) | 0.29 (0.9396) | 0.51 (0.8015) | 0.34 (0.9157) | 0.20 (0.9782) | 0.16 (0.9869) |
| H × T | **81.37 (<.0001)** | **36.07 (<.0001)** | **86.62 (<.0001)** | **42.83 (<.0001)** | **23.37 (<.0001)** | **34.04 (<.0001)** |
| H × S | 0.10 (0.9617) | 0.62 (0.6010) | 0.06 (0.9788) | 0.58 (0.6249) | 0.22 (0.8843) | 0.03 (0.9943) |
| M × C | 0.68 (0.7684) | 0.57 (0.8652) | 0.56 (0.8742) | 0.51 (0.9089) | 0.63 (0.8224) | 0.56 (0.8744) |
| M × T | **24.86 (<.0001)** | **8.84 (<.0001)** | **23.47 (<.0001)** | **9.26 (<.0001)** | **97.71 (<.0001)** | **88.99 (<.0001)** |
| M × S | 0.30 (0.9352) | 0.41 (0.8718) | 0.24 (0.9642) | 0.42 (0.8651) | 0.76 (0.6025) | **2.34 (0.0291)** |
| C × T | **2.08**<br>**(0.0014)** | **2.54**<br>**(<.0001)** | **2.53**<br>**(<.0001)** | **3.00**<br>**(<.0001)** | **8.79**<br>**(<.0001)** | **9.86**<br>**(<.0001)** |
| C × S | 0.53<br>(0.9462) | 1.23<br>(0.2228) | 0.47<br>(0.9705) | 1.17<br>(0.2736) | 1.32<br>(0.1634) | **1.98**<br>**(0.0077)** |
| T × S | **3.55**<br>**(<.0001)** | **5.47**<br>**(<.0001)** | **3.57**<br>**(<.0001)** | **5.27**<br>**(<.0001)** | 1.37<br>(0.1738) | **5.57**<br>**(<.0001)** |
| B × H × M | **69.81**<br>**(<.0001)** | **3.61**<br>**(0.0269)** | **68.79**<br>**(<.0001)** | **3.95**<br>**(0.0192)** | **51.72**<br>**(<.0001)** | **4.21**<br>**(0.0149)** |
| B × H × C | 0.27<br>(0.9492) | 1.14<br>(0.3364) | 0.20<br>(0.9767) | 1.17<br>(0.3182) | 0.12<br>(0.9938) | 0.28<br>(0.9452) |
| B × H × T | **43.28**<br>**(<.0001)** | **14.49**<br>**(<.0001)** | **46.57**<br>**(<.0001)** | **13.17**<br>**(<.0001)** | **14.02**<br>**(<.0001)** | **4.66**<br>**(0.0009)** |
| B × H × S | 0.20<br>(0.8997) | 0.23<br>(0.8736) | 0.13<br>(0.9449) | 0.15<br>(0.9283) | 0.11<br>(0.9533) | 0.05<br>(0.9870) |
| B × M × C | 0.15<br>(0.9997) | 0.31<br>(0.9882) | 0.14<br>(0.9998) | 0.33<br>(0.9846) | 0.22<br>(0.9976) | 0.14<br>(0.9997) |
| B × M × T | **13.15**<br>**(<.0001)** | **4.23**<br>**(<.0001)** | **13.80**<br>**(<.0001)** | **4.82**<br>**(<.0001)** | **17.87**<br>**(<.0001)** | **3.02**<br>**(0.0022)** |
| B × M × S | 0.16<br>(0.9876) | 0.39<br>(0.8877) | 0.14<br>(0.9914) | 0.31<br>(0.9309) | 0.10<br>(0.9967) | 0.29<br>(0.9397) |
| B × C × T | 1.06<br>(0.3872) | **1.79**<br>**(0.0099)** | 1.13<br>(0.3023) | **1.94**<br>**(0.0038)** | 1.27<br>(0.1673) | **2.40**<br>**(0.0001)** |
| B × C × S | 0.43<br>(0.9812) | 0.44<br>(0.9791) | 0.44<br>(0.9806) | 0.40<br>(0.9881) | 0.61<br>(0.8974) | 1.52<br>(0.0716) |
| B × T × S | **3.58**<br>**(<.0001)** | **2.45**<br>**(0.0035)** | **3.69**<br>**(<.0001)** | **2.59**<br>**(0.0019)** | 1.75<br>(0.0506) | **2.08**<br>**(0.0151)** |
| H × M × C | 0.10<br>(1.0000) | 0.19<br>(0.9990) | 0.10<br>(1.0000) | 0.18<br>(0.9991) | 0.05<br>(1.0000) | 0.10<br>(0.9999) |
| H × M × T | **7.76**<br>**(<.0001)** | **6.06**<br>**(<.0001)** | **7.43**<br>**(<.0001)** | **5.91**<br>**(<.0001)** | **4.83**<br>**(<.0001)** | **3.16**<br>**(0.0014)** |
| H × M × S | 0.10<br>(0.9961) | 0.10<br>(0.9965) | 0.11<br>(0.9952) | 0.08<br>(0.9983) | 0.07<br>(0.9988) | 0.20<br>(0.9776) |
| H × C × T | 0.47<br>(0.9871) | 0.27<br>(0.9998) | 0.50<br>(0.9807) | 0.28<br>(0.9998) | 0.16<br>(1.0000) | 0.17<br>(1.0000) |
| H × C × S | 0.26<br>(0.9992) | 0.25<br>(0.9995) | 0.25<br>(0.9995) | 0.20<br>(0.9999) | 0.10<br>(1.0000) | 0.07<br>(1.0000) |
| H × T × S | 0.45<br>(0.9442) | 0.27<br>(0.9937) | 0.45<br>(0.9418) | 0.28<br>(0.9921) | 0.31<br>(0.9882) | 0.07<br>(1.0000) |
| M × C × T | 0.31<br>(1.0000) | 0.25<br>(1.0000) | 0.29<br>(1.0000) | 0.23<br>(1.0000) | 0.33<br>(1.0000) | 0.32<br>(1.0000) |
| M × C × S | 0.12<br>(1.0000) | 0.20<br>(1.0000) | 0.10<br>(1.0000) | 0.18<br>(1.0000) | 0.20<br>(1.0000) | 0.25<br>(1.0000) |
| M × T × S | 0.21<br>(1.0000) | 0.59<br>(0.9455) | 0.18<br>(1.0000) | 0.57<br>(0.9515) | 0.59<br>(0.9424) | 0.81<br>(0.7272) |
| C × T × S | 0.48<br>(0.9999) | 0.87<br>(0.7710) | 0.46<br>(1.0000) | 0.84<br>(0.8231) | 1.25<br>(0.0765) | **1.63**<br>**(0.0006)** |

Table A.2. LMM results assessing the effects of *Biological Sex*, *Hand Configuration*, *Mass*, *Camera View Condition*, *MMH Task*, and *ROI Strategy* on bilateral **lateral force ($F_y$)** prediction performance metrics. Entries are *F* values (*p* values), and significant effects are highlighted in **bold** font.

| Effect | MAE (Right Hand) | MAE (Left Hand) | RMSE (Right Hand) | RMSE (Left Hand) | Peak Error (Right Hand) | Peak Error (Left Hand) |
|---|---|---|---|---|---|---|
| *Biological Sex (B)* | 1.57 (0.2195) | 0.37 (0.5463) | 1.43 (0.2399) | 0.36 (0.5532) | 3.93 (0.0557) | 0.10 (0.7510) |
| *Hand Configuration (H)* | **3040.32 (<.0001)** | **1570.57 (<.0001)** | **3371.78 (<.0001)** | **1652.88 (<.0001)** | **3287.03 (<.0001)** | **840.23 (<.0001)** |
| *Mass (M)* | **1474.30 (<.0001)** | **595.66 (<.0001)** | **1723.87 (<.0001)** | **681.23 (<.0001)** | **194.64 (<.0001)** | **109.90 (<.0001)** |
| *Camera View Condition (C)* | 2.06 (0.0543) | 0.60 (0.7308) | **5.25 (<.0001)** | 0.43 (0.8600) | **12.62 (<.0001)** | **3.11 (0.0048)** |
| *MMH Task (T)* | **4326.06 (<.0001)** | **2245.52 (<.0001)** | **4691.98 (<.0001)** | **2452.49 (<.0001)** | **672.43 (<.0001)** | **578.96 (<.0001)** |
| *ROI Strategy (S)* | 2.57 (0.0526) | 1.16 (0.3227) | **2.86 (0.0352)** | 1.15 (0.3277) | 1.26 (0.2845) | 0.89 (0.4448) |
| B × H | **329.06 (<.0001)** | 3.82 (0.0507) | **361.09 (<.0001)** | **5.50 (0.0190)** | **204.98 (<.0001)** | 0.33 (0.5676) |
| B × M | **4.89 (0.0076)** | **23.69 (<.0001)** | **5.08 (0.0062)** | **25.94 (<.0001)** | **29.09 (<.0001)** | **64.37 (<.0001)** |
| B × C | 0.95 (0.4582) | 0.42 (0.8686) | 1.09 (0.3642) | 0.42 (0.8651) | **2.40 (0.0256)** | 0.65 (0.6930) |
| B × T | **117.22 (<.0001)** | **86.23 (<.0001)** | **120.03 (<.0001)** | **93.89 (<.0001)** | **159.77 (<.0001)** | **36.25 (<.0001)** |
| B × S | **3.60 (0.0129)** | 0.58 (0.6268) | **4.32 (0.0047)** | 0.57 (0.6380) | 2.02 (0.1088) | 0.47 (0.7046) |
| H × M | **23.07 (<.0001)** | **127.42 (<.0001)** | **24.85 (<.0001)** | **132.63 (<.0001)** | **47.29 (<.0001)** | **116.01 (<.0001)** |
| H × C | 0.58 (0.7441) | 0.18 (0.9835) | 0.54 (0.7786) | 0.14 (0.9908) | 1.30 (0.2522) | 0.64 (0.6957) |
| H × T | **69.64 (<.0001)** | **44.09 (<.0001)** | **67.76 (<.0001)** | **34.53 (<.0001)** | **204.60 (<.0001)** | **46.07 (<.0001)** |
| H × S | 0.70 (0.5535) | 0.08 (0.9696) | 0.58 (0.6290) | 0.11 (0.9548) | 0.91 (0.4334) | 0.34 (0.7952) |
| M × C | 0.13 (0.9998) | 0.16 (0.9995) | 0.09 (1.0000) | 0.16 (0.9995) | 1.11 (0.3456) | 0.54 (0.8934) |
| M × T | **77.81 (<.0001)** | **24.69 (<.0001)** | **85.94 (<.0001)** | **25.04 (<.0001)** | **15.44 (<.0001)** | **9.96 (<.0001)** |
| M × S | 0.05 (0.9995) | 0.95 (0.4604) | 0.09 (0.9973) | 0.91 (0.4828) | 0.17 (0.9837) | 0.67 (0.6763) |
| C × T | 0.53<br>(0.9694) | 0.47<br>(0.9878) | 0.53<br>(0.9688) | 0.50<br>(0.9797) | 1.49<br>(0.0587) | 1.08<br>(0.3631) |
| C × S | 1.04<br>(0.4049) | 0.52<br>(0.9520) | 1.08<br>(0.3631) | 0.57<br>(0.9234) | 0.51<br>(0.9547) | 0.48<br>(0.9672) |
| T × S | 0.82<br>(0.6260) | 0.90<br>(0.5438) | 0.80<br>(0.6472) | 1.00<br>(0.4438) | **1.80**<br>**(0.0423)** | **2.52**<br>**(0.0026)** |
| B × H × M | **11.46**<br>**(<.0001)** | **9.36**<br>**(<.0001)** | **7.60**<br>**(0.0005)** | **8.25**<br>**(0.0003)** | **5.94**<br>**(0.0026)** | **7.93**<br>**(0.0004)** |
| B × H × C | 0.32<br>(0.9271) | 0.09<br>(0.9977) | 0.25<br>(0.9581) | 0.09<br>(0.9973) | 0.33<br>(0.9206) | 0.11<br>(0.9958) |
| B × H × T | **48.65**<br>**(<.0001)** | 2.30<br>(0.0565) | **43.95**<br>**(<.0001)** | 2.01<br>(0.0902) | **23.67**<br>**(<.0001)** | **4.81**<br>**(0.0007)** |
| B × H × S | 2.33<br>(0.0723) | 0.17<br>(0.9139) | 2.30<br>(0.0747) | 0.13<br>(0.9414) | 0.02<br>(0.9972) | 0.03<br>(0.9935) |
| B × M × C | 0.31<br>(0.9872) | 0.21<br>(0.9980) | 0.33<br>(0.9846) | 0.18<br>(0.9991) | 0.24<br>(0.9961) | 0.08<br>(1.0000) |
| B × M × T | **7.63**<br>**(<.0001)** | **2.13**<br>**(0.0295)** | **7.63**<br>**(<.0001)** | **2.47**<br>**(0.0112)** | **12.76**<br>**(<.0001)** | **3.11**<br>**(0.0016)** |
| B × M × S | 0.23<br>(0.9681) | 0.12<br>(0.9936) | 0.24<br>(0.9619) | 0.13<br>(0.9925) | 0.55<br>(0.7667) | 0.10<br>(0.9959) |
| B × C × T | 1.31<br>(0.1448) | 0.65<br>(0.9004) | 1.29<br>(0.1585) | 0.64<br>(0.9069) | **1.73**<br>**(0.0148)** | 1.01<br>(0.4505) |
| B × C × S | 0.64<br>(0.8734) | 0.43<br>(0.9814) | 0.65<br>(0.8663) | 0.42<br>(0.9841) | 0.39<br>(0.9903) | 0.57<br>(0.9248) |
| B × T × S | 1.40<br>(0.1556) | 1.38<br>(0.1665) | 1.59<br>(0.0881) | 1.28<br>(0.2220) | 1.03<br>(0.4148) | 0.67<br>(0.7833) |
| H × M × C | 0.14<br>(0.9997) | 0.06<br>(1.0000) | 0.13<br>(0.9998) | 0.06<br>(1.0000) | 0.08<br>(1.0000) | 0.09<br>(1.0000) |
| H × M × T | **16.52**<br>**(<.0001)** | **11.54**<br>**(<.0001)** | **15.32**<br>**(<.0001)** | **13.52**<br>**(<.0001)** | **7.04**<br>**(<.0001)** | **12.80**<br>**(<.0001)** |
| H × M × S | 0.11<br>(0.9949) | 0.10<br>(0.9966) | 0.10<br>(0.9963) | 0.10<br>(0.9963) | 0.08<br>(0.9984) | 0.04<br>(0.9998) |
| H × C × T | 0.70<br>(0.8579) | 0.18<br>(1.0000) | 0.70<br>(0.8554) | 0.15<br>(1.0000) | 0.30<br>(0.9997) | 0.13<br>(1.0000) |
| H × C × S | 0.31<br>(0.9974) | 0.18<br>(0.9999) | 0.32<br>(0.9970) | 0.18<br>(1.0000) | 0.10<br>(1.0000) | 0.06<br>(1.0000) |
| H × T × S | 0.25<br>(0.9951) | 0.49<br>(0.9230) | 0.27<br>(0.9932) | 0.46<br>(0.9373) | 0.24<br>(0.9962) | 0.20<br>(0.9986) |
| M × C × T | 0.37<br>(1.0000) | 0.13<br>(1.0000) | 0.35<br>(1.0000) | 0.14<br>(1.0000) | 0.35<br>(1.0000) | 0.18<br>(1.0000) |
| M × C × S | 0.12<br>(1.0000) | 0.06<br>(1.0000) | 0.12<br>(1.0000) | 0.06<br>(1.0000) | 0.11<br>(1.0000) | 0.08<br>(1.0000) |
| M × T × S | 0.22<br>(1.0000) | 0.18<br>(1.0000) | 0.19<br>(1.0000) | 0.16<br>(1.0000) | 0.31<br>(0.9995) | 0.14<br>(1.0000) |
| C × T × S | 0.44<br>(1.0000) | 0.40<br>(1.0000) | 0.42<br>(1.0000) | 0.38<br>(1.0000) | 0.73<br>(0.9567) | 0.51<br>(0.9998) |

Table A.3. LMM results assessing the effects of *Biological Sex*, *Hand Configuration*, *Mass*, *Camera View Condition*, *MMH Task*, and *ROI Strategy* on bilateral **vertical force ($F_z$)** prediction performance metrics. Entries are *F* values (*p* values), and significant effects are highlighted in **bold** font.

| Effect | MAE (Right Hand) | MAE (Left Hand) | RMSE (Right Hand) | RMSE (Left Hand) | Peak Error (Right Hand) | Peak Error (Left Hand) |
|---|---|---|---|---|---|---|
| *Biological Sex (B)* | **4.57 (0.0400)** | 4.06 (0.0521) | **4.69 (0.0376)** | **4.55 (0.0404)** | 2.02 (0.1643) | 1.21 (0.2796) |
| *Hand Configuration (H)* | **32.74 (<.0001)** | **96.96 (<.0001)** | **29.86 (<.0001)** | **101.01 (<.0001)** | **117.04 (<.0001)** | **29.90 (<.0001)** |
| *Mass (M)* | **1743.82 (<.0001)** | **1843.24 (<.0001)** | **2032.50 (<.0001)** | **2111.07 (<.0001)** | **299.89 (<.0001)** | **365.72 (<.0001)** |
| *Camera View Condition (C)* | **11.40 (<.0001)** | **13.02 (<.0001)** | **10.42 (<.0001)** | **9.13 (<.0001)** | **40.67 (<.0001)** | **39.82 (<.0001)** |
| *MMH Task (T)* | **477.71 (<.0001)** | **821.42 (<.0001)** | **385.21 (<.0001)** | **715.55 (<.0001)** | **80.30 (<.0001)** | **181.87 (<.0001)** |
| *ROI Strategy (S)* | **43.20 (<.0001)** | **15.70 (<.0001)** | **45.06 (<.0001)** | **17.35 (<.0001)** | **8.55 (<.0001)** | **10.02 (<.0001)** |
| B × H | **167.51 (<.0001)** | **73.47 (<.0001)** | **172.53 (<.0001)** | **84.33 (<.0001)** | **8.68 (0.0032)** | **6.58 (0.0103)** |
| B × M | **9.58 (<.0001)** | **3.99 (0.0184)** | **7.66 (0.0005)** | 2.91 (0.0548) | **4.84 (0.0079)** | 1.68 (0.1867) |
| B × C | 0.72 (0.6341) | 0.86 (0.5239) | 0.74 (0.6165) | 0.80 (0.5720) | **2.87 (0.0085)** | **3.55 (0.0016)** |
| B × T | **300.58 (<.0001)** | **324.57 (<.0001)** | **333.79 (<.0001)** | **344.53 (<.0001)** | **19.69 (<.0001)** | **129.67 (<.0001)** |
| B × S | **10.83 (<.0001)** | **11.71 (<.0001)** | **13.09 (<.0001)** | **12.52 (<.0001)** | 0.35 (0.7861) | 1.88 (0.1305) |
| H × M | **23.56 (<.0001)** | **39.92 (<.0001)** | **22.93 (<.0001)** | **32.90 (<.0001)** | **25.42 (<.0001)** | 1.86 (0.1555) |
| H × C | 0.20 (0.9754) | 0.48 (0.8216) | 0.20 (0.9770) | 0.44 (0.8540) | 0.36 (0.9042) | 0.10 (0.9968) |
| H × T | **11.48 (<.0001)** | **12.06 (<.0001)** | **11.68 (<.0001)** | **13.71 (<.0001)** | **26.01 (<.0001)** | **23.95 (<.0001)** |
| H × S | 0.80 (0.4945) | 0.21 (0.8894) | 0.63 (0.5946) | 0.12 (0.9463) | 0.24 (0.8661) | 0.02 (0.9957) |
| M × C | 0.39 (0.9688) | 0.55 (0.8820) | 0.37 (0.9727) | 0.38 (0.9702) | 1.25 (0.2399) | 1.36 (0.1778) |
| M × T | **106.85 (<.0001)** | **113.70 (<.0001)** | **116.41 (<.0001)** | **122.73 (<.0001)** | **15.84 (<.0001)** | **14.28 (<.0001)** |
| M × S | 0.22 (0.9695) | 0.43 (0.8584) | 0.17 (0.9854) | 0.33 (0.9192) | **2.74 (0.0115)** | 1.67 (0.1238) |
| C × T | **1.85**<br>**(0.0071)** | **2.38**<br>**(0.0002)** | **1.93**<br>**(0.0041)** | **2.47**<br>**(<.0001)** | **3.05**<br>**(<.0001)** | **2.24**<br>**(0.0005)** |
| C × S | 0.92<br>(0.5570) | 1.11<br>(0.3368) | 0.90<br>(0.5829) | 1.08<br>(0.3664) | **1.71**<br>**(0.0303)** | 1.22<br>(0.2320) |
| T × S | **2.39**<br>**(0.0043)** | **3.84**<br>**(<.0001)** | **2.23**<br>**(0.0085)** | **3.89**<br>**(<.0001)** | 1.23<br>(0.2564) | 1.39<br>(0.1642) |
| B × H × M | **41.92**<br>**(<.0001)** | **52.08**<br>**(<.0001)** | **43.91**<br>**(<.0001)** | **55.31**<br>**(<.0001)** | 0.79<br>(0.4528) | **9.38**<br>**(<.0001)** |
| B × H × C | 0.36<br>(0.9064) | 0.18<br>(0.9820) | 0.29<br>(0.9421) | 0.16<br>(0.9872) | 0.04<br>(0.9998) | 0.09<br>(0.9977) |
| B × H × T | **5.18**<br>**(0.0004)** | **10.12**<br>**(<.0001)** | **7.54**<br>**(<.0001)** | **12.70**<br>**(<.0001)** | **3.56**<br>**(0.0066)** | 0.97<br>(0.4222) |
| B × H × S | 0.42<br>(0.7379) | 0.05<br>(0.9840) | 0.35<br>(0.7907) | 0.06<br>(0.9803) | 0.25<br>(0.8622) | 0.35<br>(0.7918) |
| B × M × C | 0.64<br>(0.8053) | 0.18<br>(0.9991) | 0.65<br>(0.7974) | 0.21<br>(0.9983) | 0.20<br>(0.9985) | 0.10<br>(1.0000) |
| B × M × T | **13.03**<br>**(<.0001)** | **10.79**<br>**(<.0001)** | **13.95**<br>**(<.0001)** | **11.41**<br>**(<.0001)** | **6.29**<br>**(<.0001)** | **5.18**<br>**(<.0001)** |
| B × M × S | 0.24<br>(0.9624) | 0.56<br>(0.7640) | 0.27<br>(0.9496) | 0.54<br>(0.7765) | 0.14<br>(0.9910) | 0.04<br>(0.9998) |
| B × C × T | 0.98<br>(0.4953) | 1.08<br>(0.3557) | 1.01<br>(0.4487) | 1.10<br>(0.3373) | 1.46<br>(0.0674) | **2.26**<br>**(0.0004)** |
| B × C × S | 0.75<br>(0.7617) | 1.56<br>(0.0623) | 0.74<br>(0.7734) | 1.56<br>(0.0615) | 0.52<br>(0.9507) | 0.76<br>(0.7491) |
| B × T × S | 1.46<br>(0.1302) | 1.25<br>(0.2394) | 1.47<br>(0.1263) | 1.17<br>(0.3010) | 1.71<br>(0.0571) | 1.50<br>(0.1164) |
| H × M × C | 0.28<br>(0.9919) | 0.13<br>(0.9998) | 0.27<br>(0.9940) | 0.14<br>(0.9998) | 0.05<br>(1.0000) | 0.06<br>(1.0000) |
| H × M × T | **10.70**<br>**(<.0001)** | **16.06**<br>**(<.0001)** | **10.03**<br>**(<.0001)** | **14.41**<br>**(<.0001)** | **3.37**<br>**(0.0007)** | **7.20**<br>**(<.0001)** |
| H × M × S | 0.31<br>(0.9344) | 0.18<br>(0.9814) | 0.28<br>(0.9458) | 0.16<br>(0.9862) | 0.15<br>(0.9884) | 0.13<br>(0.9930) |
| H × C × T | 0.36<br>(0.9982) | 0.39<br>(0.9967) | 0.32<br>(0.9994) | 0.36<br>(0.9984) | 0.13<br>(1.0000) | 0.26<br>(0.9999) |
| H × C × S | 0.21<br>(0.9999) | 0.17<br>(1.0000) | 0.20<br>(0.9999) | 0.15<br>(1.0000) | 0.05<br>(1.0000) | 0.05<br>(1.0000) |
| H × T × S | 0.30<br>(0.9893) | 0.23<br>(0.9970) | 0.32<br>(0.9869) | 0.20<br>(0.9984) | 0.12<br>(0.9999) | 0.07<br>(1.0000) |
| M × C × T | 0.30<br>(1.0000) | 0.39<br>(1.0000) | 0.30<br>(1.0000) | 0.36<br>(1.0000) | 0.40<br>(0.9999) | 0.42<br>(0.9999) |
| M × C × S | 0.22<br>(1.0000) | 0.19<br>(1.0000) | 0.19<br>(1.0000) | 0.18<br>(1.0000) | 0.19<br>(1.0000) | 0.18<br>(1.0000) |
| M × T × S | 0.34<br>(0.9989) | 0.29<br>(0.9997) | 0.35<br>(0.9985) | 0.25<br>(0.9999) | 0.37<br>(0.9978) | 0.18<br>(1.0000) |
| C × T × S | 0.80<br>(0.8921) | 0.62<br>(0.9957) | 0.80<br>(0.8907) | 0.62<br>(0.9954) | 0.95<br>(0.6003) | 0.97<br>(0.5558) |

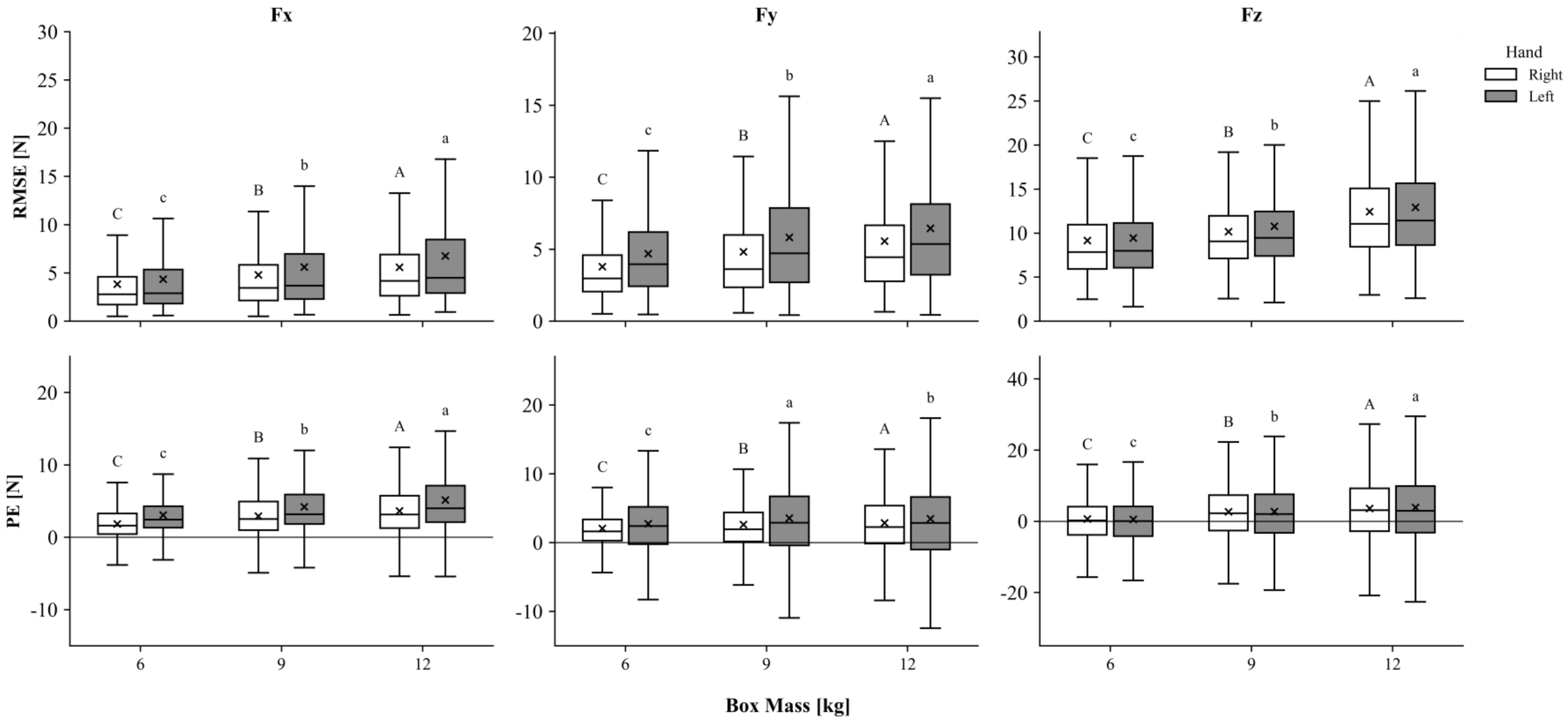

Figure A.5. Effect of *Mass* on RMSE (top row) and PE (bottom row) of $F_x$ (left), $F_y$ (middle), and $F_z$ (right), for the right and left hands (white and gray, respectively). Based on *post hoc* pairwise comparisons, *Mass* levels that do not share a common letter are significantly different within a hand; right-hand comparisons are shown in uppercase letters and left-hand comparisons in lowercase letters.

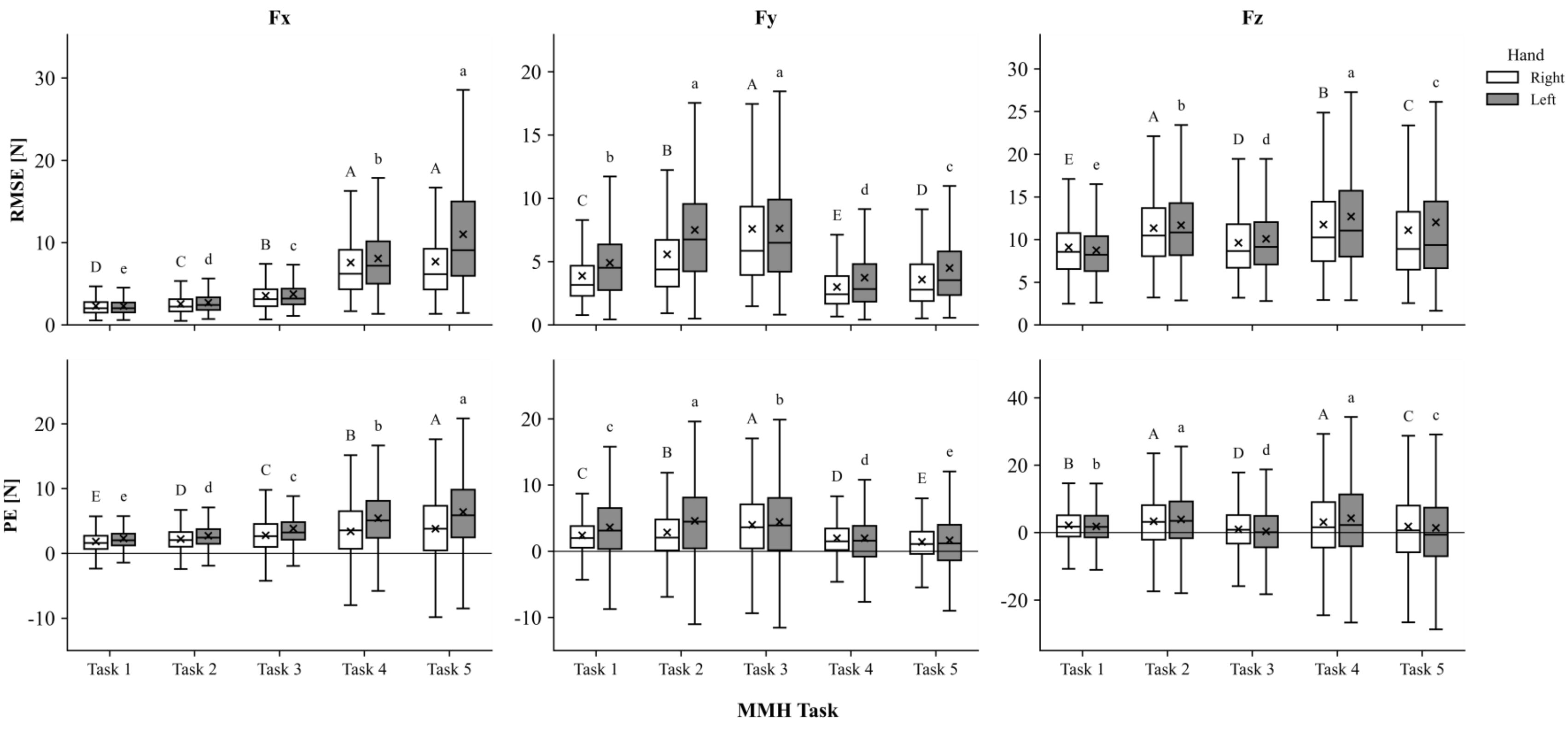

Figure A.6. Effect of *MMH Task* on RMSE (top row) and PE (bottom row) of $F_x$ (left), $F_y$ (middle), and $F_z$ (right), for the right and left hands (white and gray, respectively). Based on *post hoc* pairwise comparisons, *MMH Tasks* that do not share a common letter are significantly different within a hand; right-hand comparisons are shown in uppercase letters and left-hand comparisons in lowercase letters.

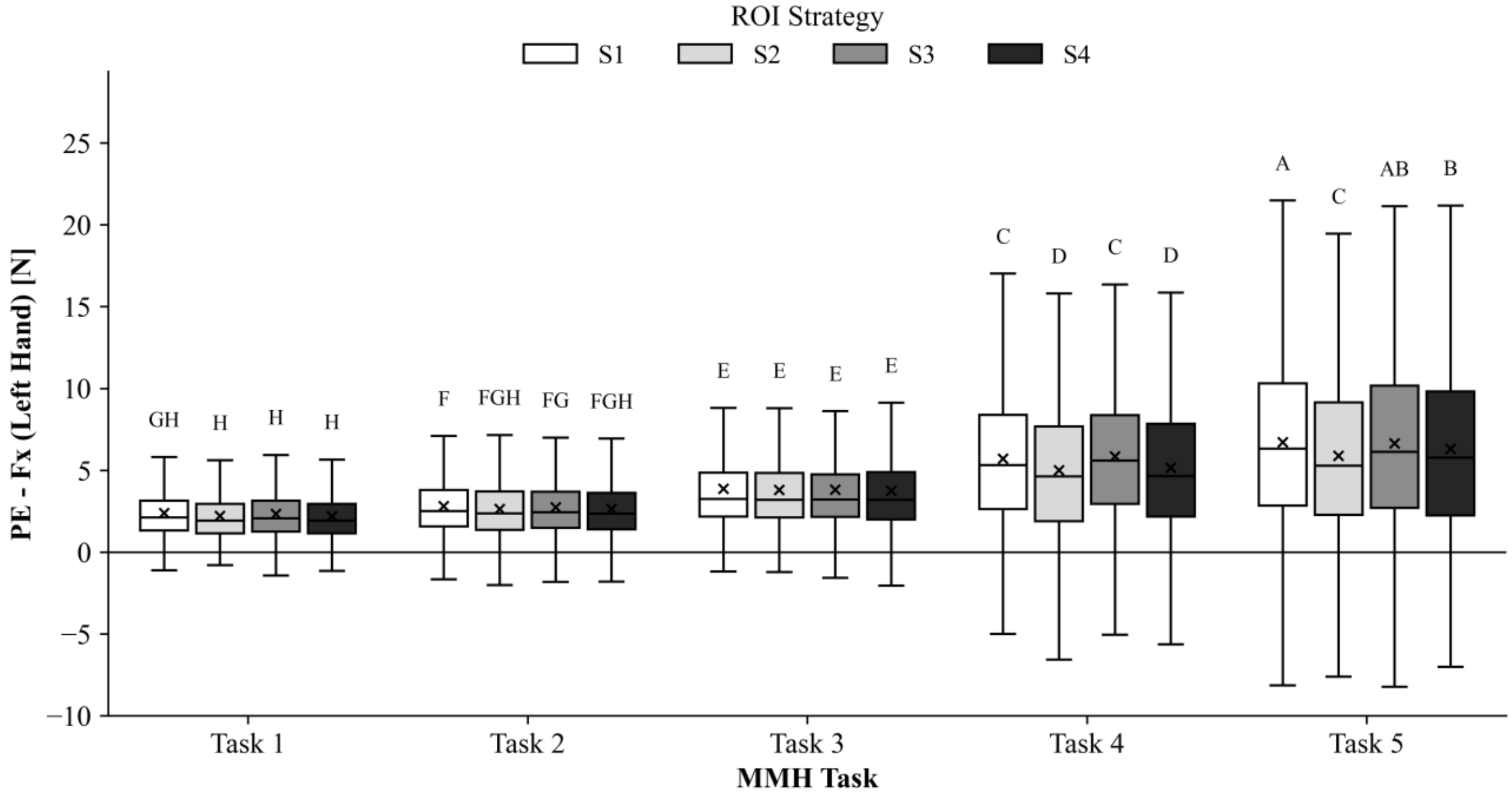

Figure A.7. *ROI Strategy* × *MMH Task* interaction effect on $F_x$ PE for the left hand. Based on *post hoc* pairwise comparisons, boxes that do not share a common letter are significantly different.

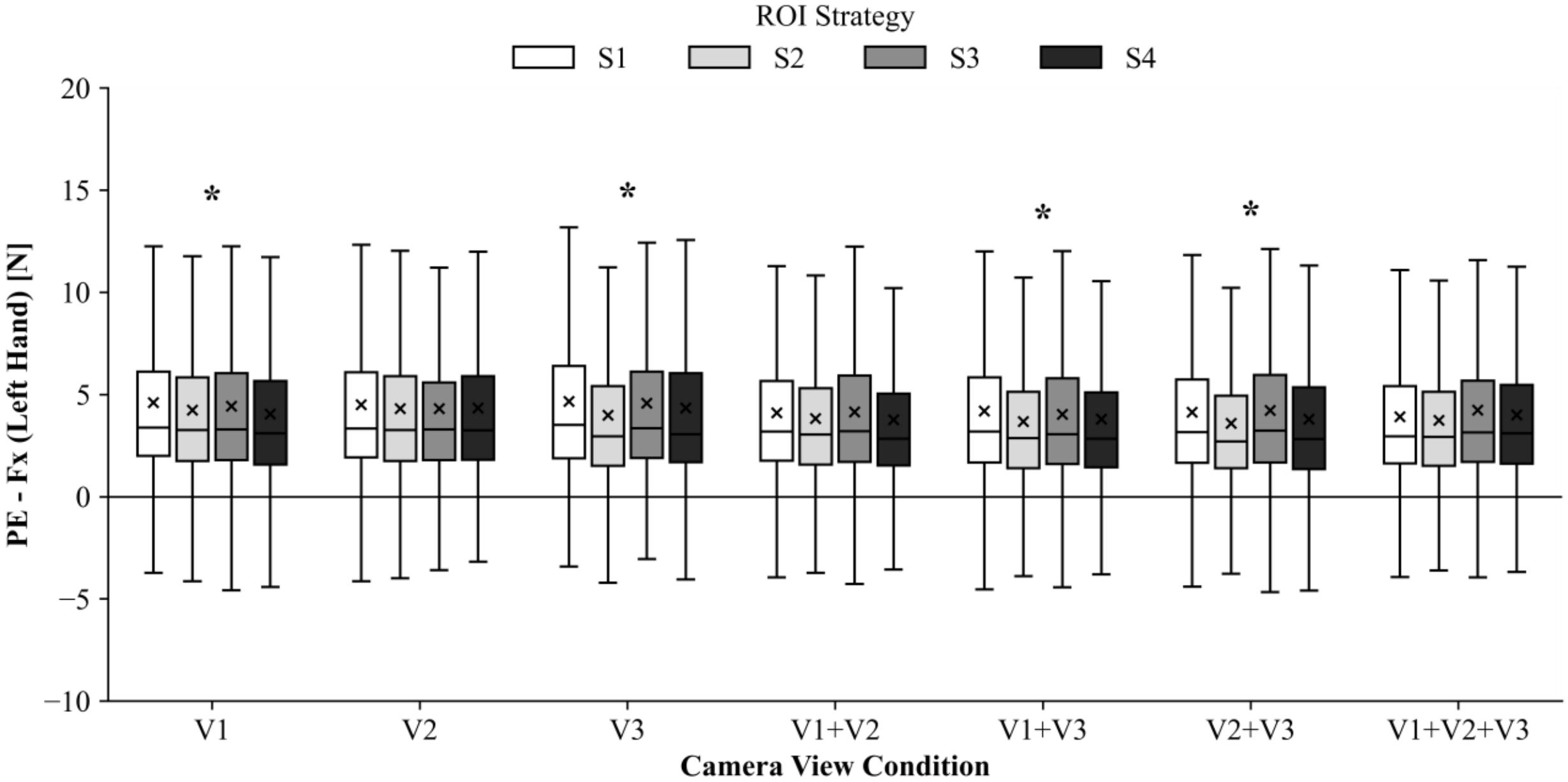

Figure A.8. *Camera View Condition* × *ROI Strategy* interaction effect on $F_x$ PE for the left hand. Asterisks denote a significant effect of *ROI Strategy* within a specific *Camera View Condition*.

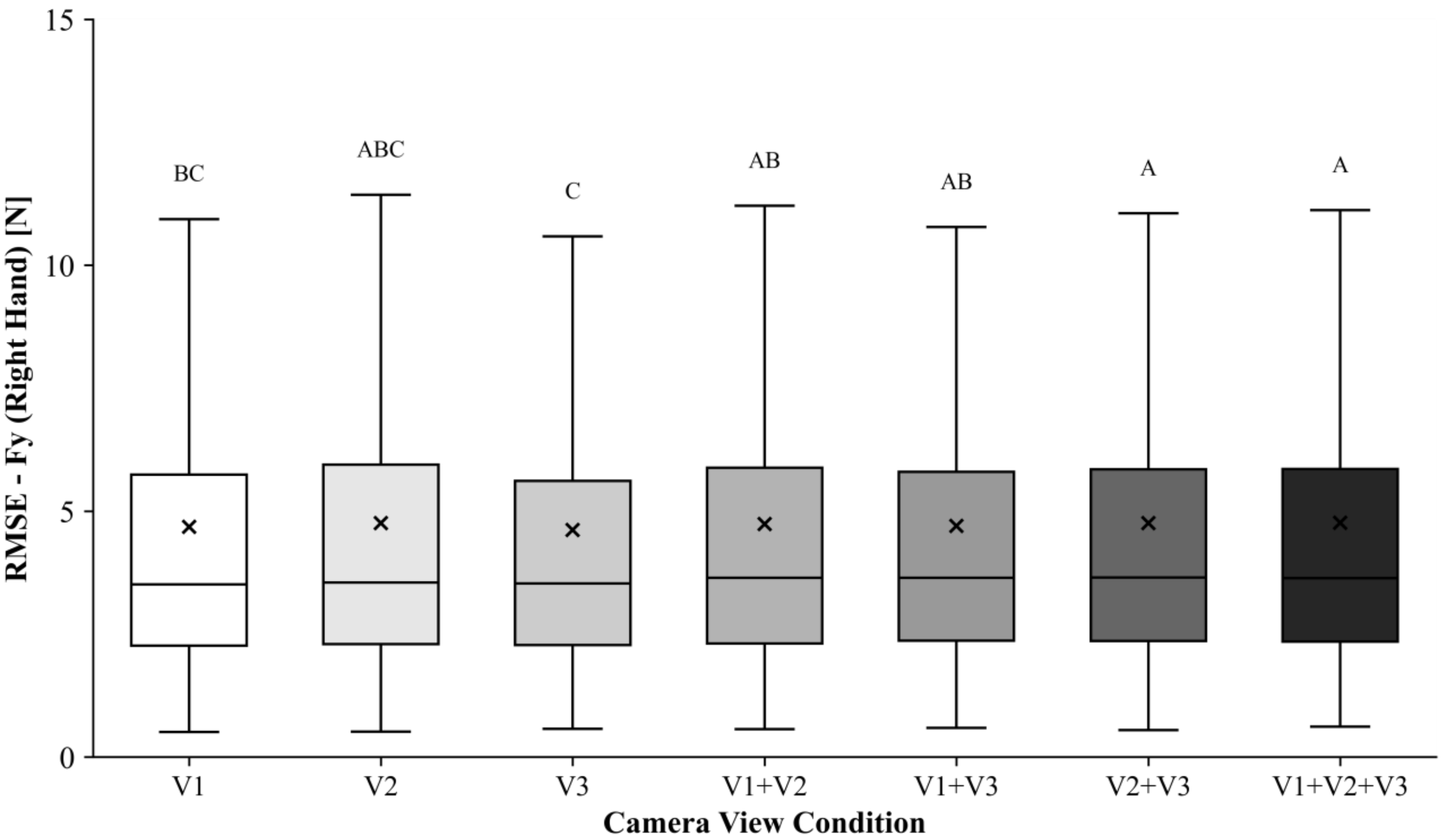

Figure A.9. Effect of *Camera View Condition* on $F_y$ RMSE for the right hand. Based on *post hoc* pairwise comparisons, boxes that do not share a common letter are significantly different.

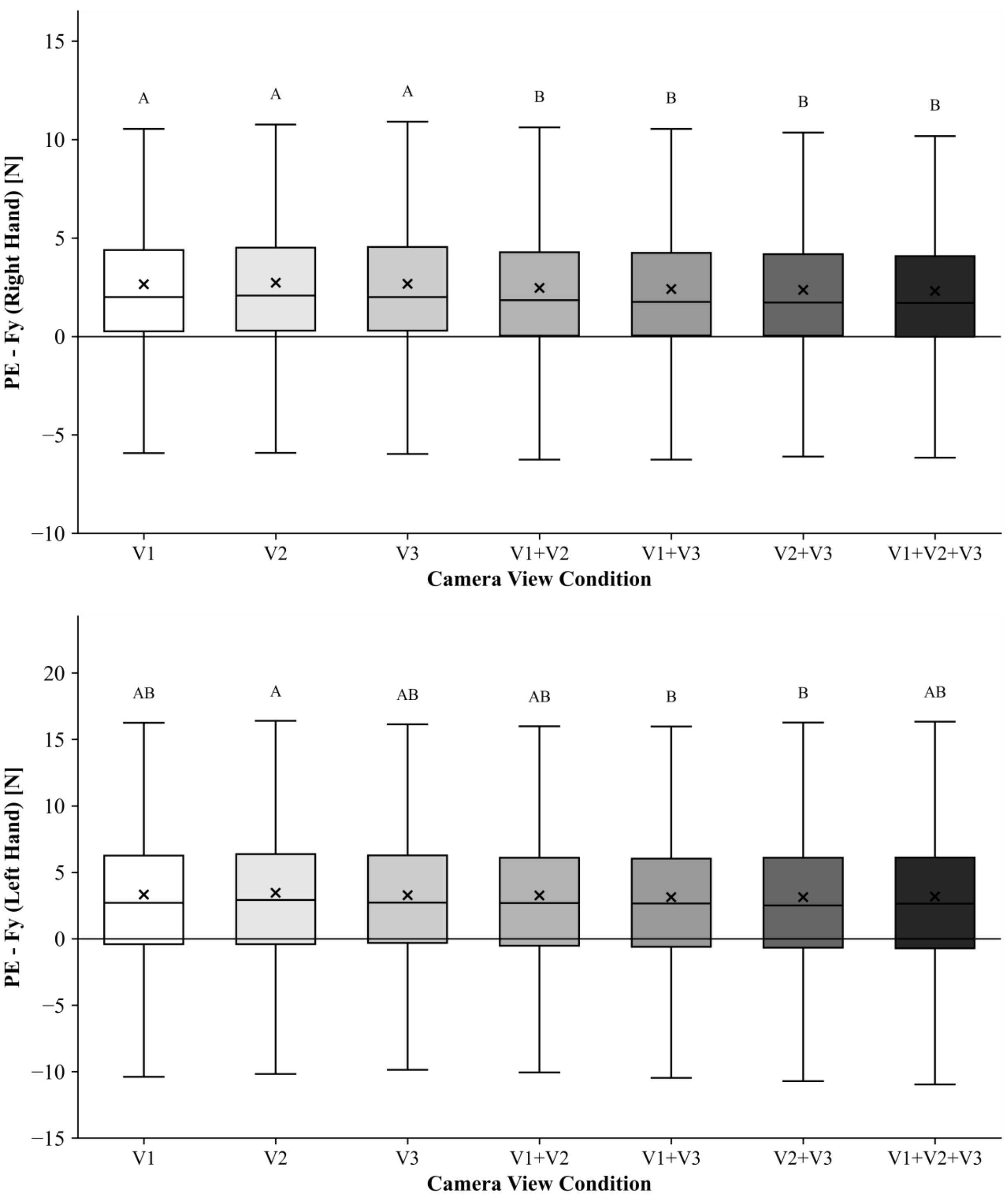

Figure A.10. Effect of *Camera View Condition* on $F_y$ PE for the right (top) and left (bottom) hands. Based on *post hoc* pairwise comparisons, boxes that do not share a common letter are significantly different.

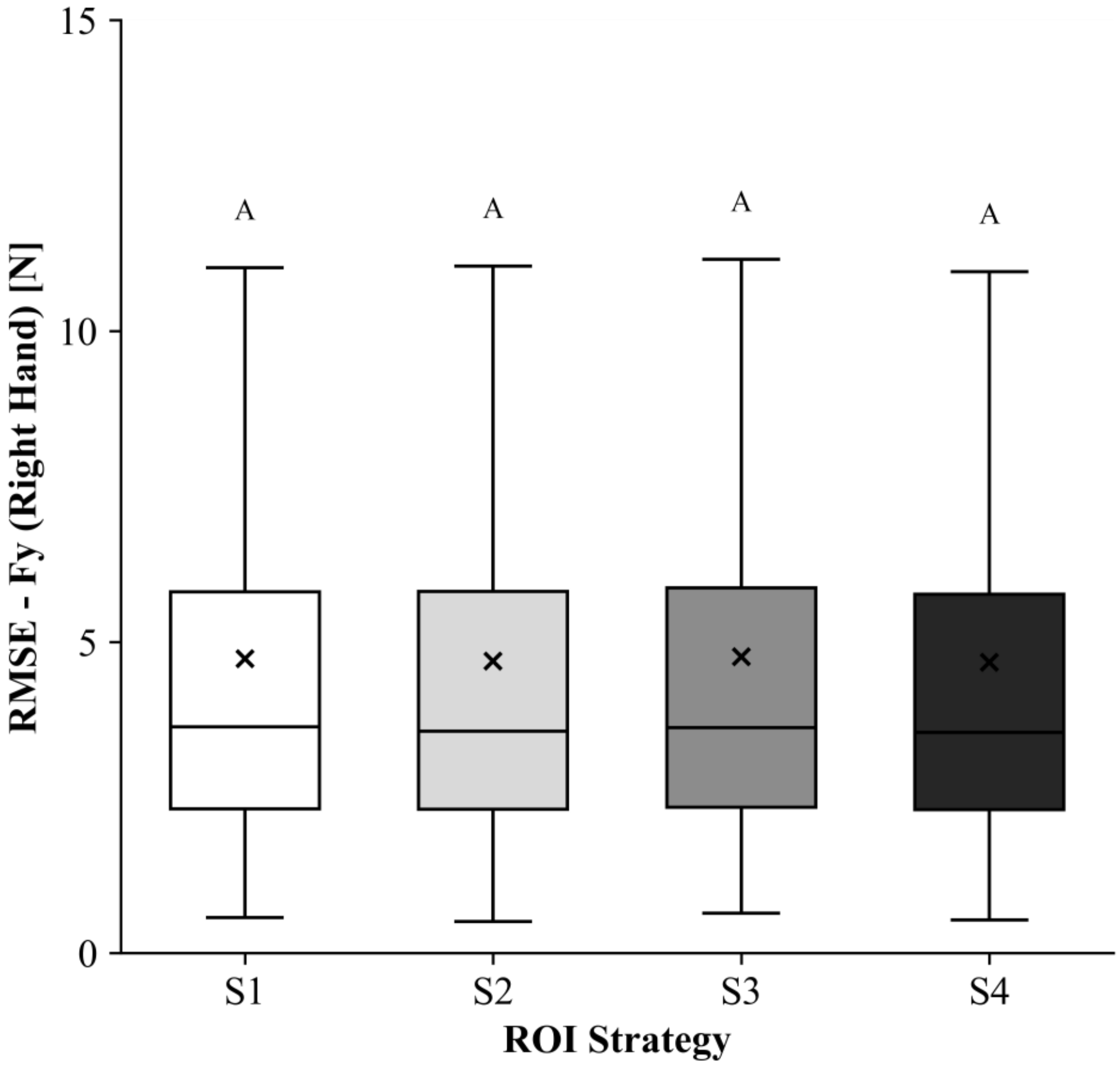

Figure A.11. Effect of *ROI Strategy* on $F_y$ RMSE for the right hand. Based on *post hoc* pairwise comparisons, boxes that do not share a common letter are significantly different.

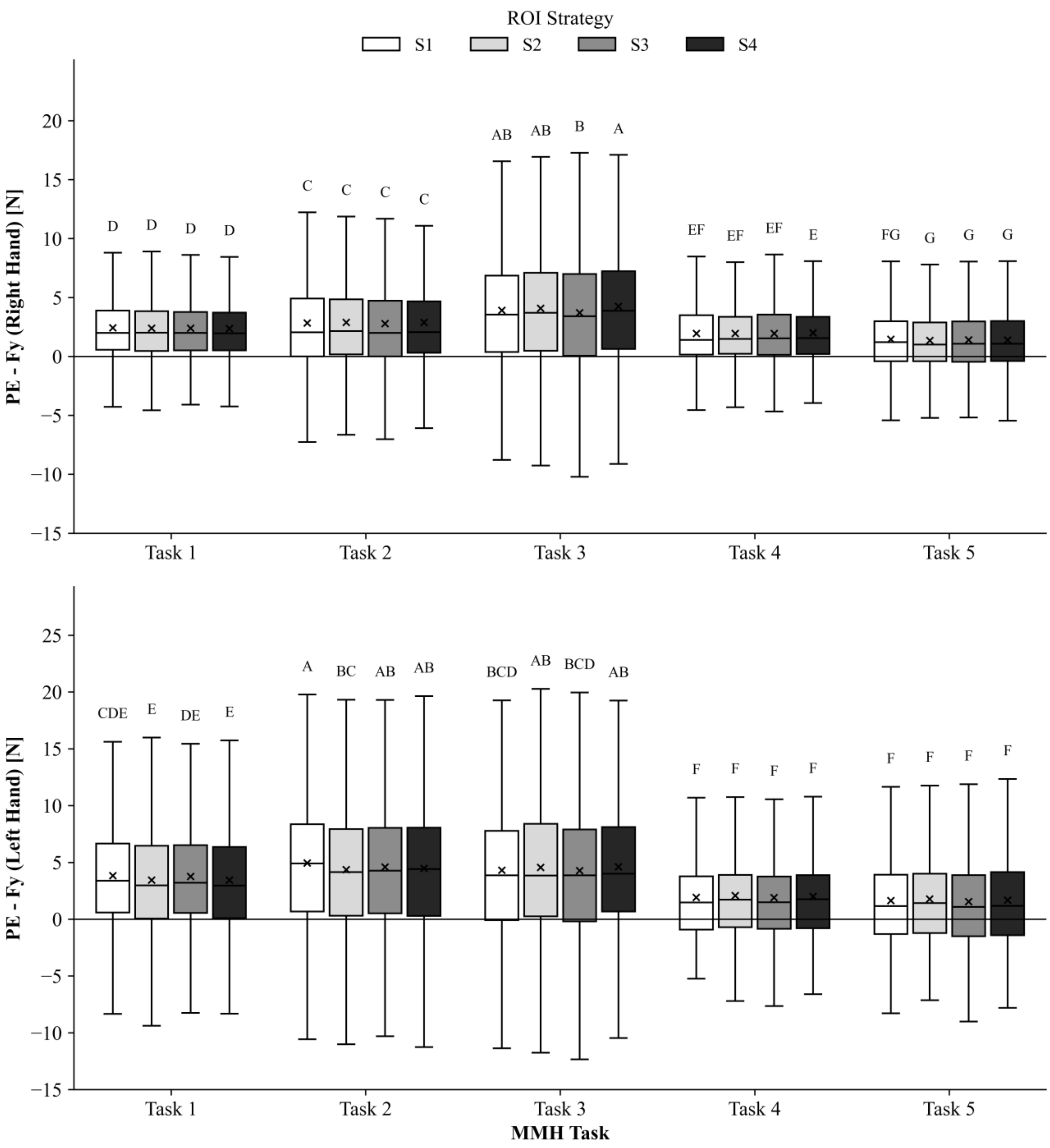

Figure A.12. *ROI Strategy* × *MMH Task* interaction effect on $F_y$ PE for the right (top) and left (bottom) hands. Based on *post hoc* pairwise comparisons, boxes that do not share a common letter are significantly different.

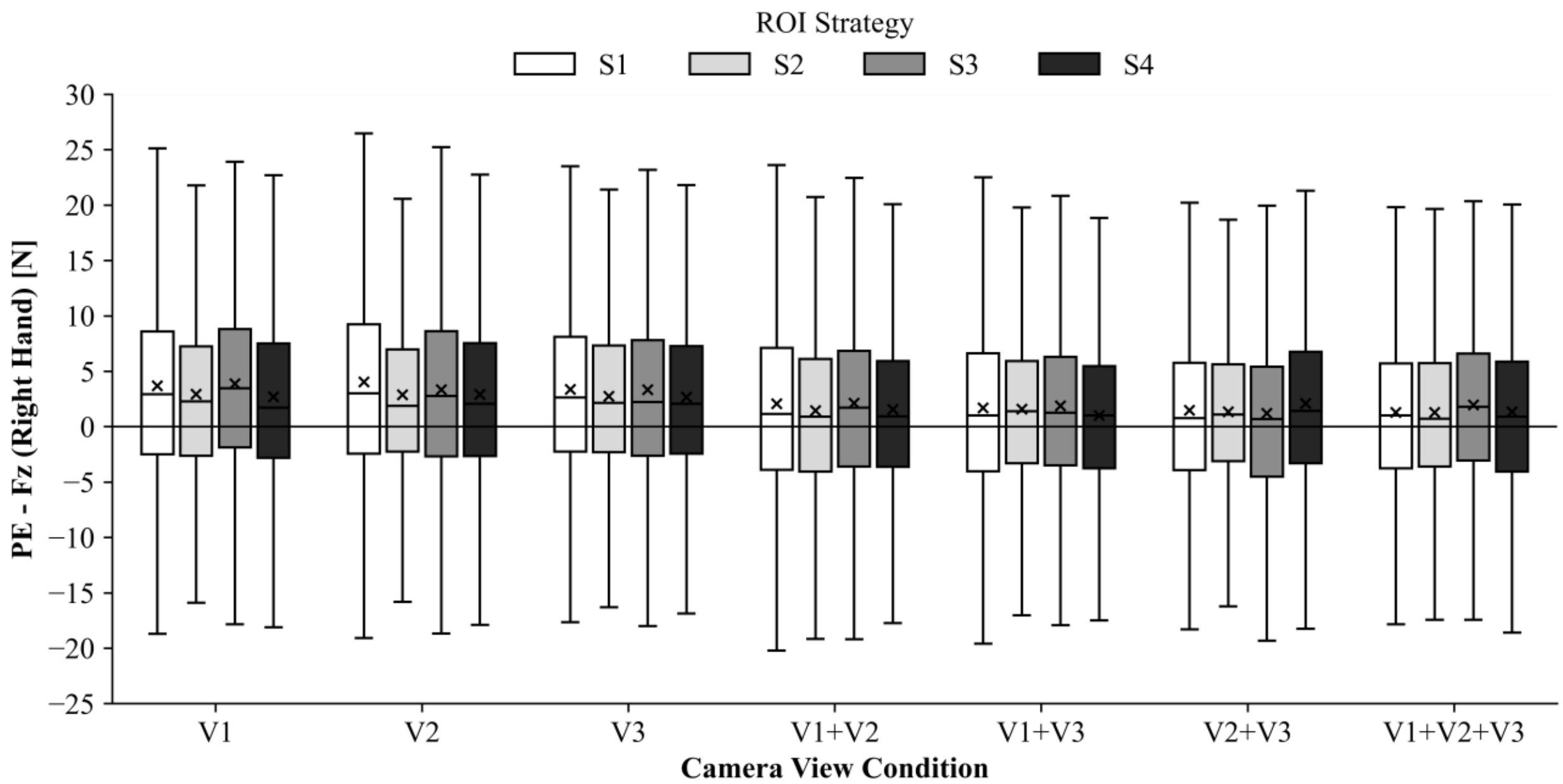

Figure A.13. *Camera View Condition* × *ROI Strategy* interaction effect on $F_z$ PE for the right hand.